\documentclass[10pt,journal,compsoc]{IEEEtran}
\usepackage{color}
\usepackage{xspace}
\usepackage{amsfonts}
\usepackage{amsmath}
\usepackage{amssymb}
\usepackage{bm}
\usepackage{ragged2e}
\usepackage{booktabs}
\usepackage{multirow}
\usepackage{graphicx}
\usepackage{threeparttable}
\usepackage{makecell}
\usepackage{array}

\usepackage{epsfig}
\usepackage{arydshln, collcell}
\usepackage{algorithm}
\usepackage{algorithmic}
\newcolumntype{?}[1]{!{\vrule width #1}}
\usepackage{graphbox}

\usepackage[caption=false]{subfig}
\makeatletter
\DeclareRobustCommand\onedot{\futurelet\@let@token\@onedot}
\def\@onedot{\ifx\@let@token.\else.\null\fi\xspace}

\def\ie{\emph{i.e}\onedot} 
 
\def\etc{\emph{etc}\onedot} 
 
\def\etal{\emph{et al}\onedot}

\makeatother

\usepackage[pagebackref=true,breaklinks=true,letterpaper=true,bookmarks=false]{hyperref} 

\ifCLASSOPTIONcompsoc
  \usepackage[nocompress]{cite}
\else
  \usepackage{cite}
\fi
\ifCLASSINFOpdf
\else
\fi
\begin{document}
%
\title{EventHDR: from Event to High-Speed HDR Videos and Beyond}
%
%
%
%

\author{
	Yunhao~Zou,~\IEEEmembership{Student Member,~IEEE,}
	Ying~Fu,~\IEEEmembership{Senior Member,~IEEE,}
	Tsuyoshi~Takatani,~\IEEEmembership{Member,~IEEE}
	and ~Yinqiang~Zheng,~\IEEEmembership{Senior Member,~IEEE}
	\IEEEcompsocitemizethanks{
		\IEEEcompsocthanksitem Yunhao Zou and Ying Fu are with MIIT Key Laboratory of Complex-field Intelligent Sensing, Beijing Institute of Technology, Beijing, China, and School of Computer Science and Technology, Beijing Institute of Technology, Beijing, China. (e-mail: zouyunhao@bit.edu.cn; fuying@bit.edu.cn).
		\IEEEcompsocthanksitem Tsuyoshi Takatani is with the Institute of Systems and Information Engineering, the University of Tsukuba, Tsukuba, Japan. (email: takatani@iit.tsukuba.ac.jp).
		\IEEEcompsocthanksitem Yinqiang Zheng is with the Next Generation Artiﬁcial Intelligence Research Center, the University of Tokyo, Tokyo 113-8656, Japan. (e-mail: yqzheng@ai.u-tokyo.ac.jp).
		\IEEEcompsocthanksitem Corresponding author: Ying Fu
	}
		\thanks{
		This work was supported by the National Natural Science Foundation of China (62331006, 62171038, and 62088101), the Fundamental Research Funds for the Central Universities, and JSPS KAKENHI (24K22318, 22H00529).}
}

%
%

\markboth{IEEE TRANSACTIONS ON PATTERN ANALYSIS AND MACHINE INTELLIGENCE }%
{Shell \MakeLowercase{\textit{et al.}}: Bare Demo of IEEEtran.cls for Computer Society Journals}
%



\IEEEtitleabstractindextext{%
\begin{abstract}
\justifying
Event cameras are innovative neuromorphic sensors that asynchronously capture the scene dynamics. Due to the event-triggering mechanism, such cameras record event streams with much shorter response latency and higher intensity sensitivity compared to conventional cameras.  On the basis of these features, previous works have attempted to reconstruct high dynamic range (HDR) videos from events, but have either suffered from unrealistic artifacts or failed to provide sufficiently high frame rates. In this paper, we present a recurrent convolutional neural network that reconstruct high-speed HDR videos from event sequences, with a key frame guidance to prevent potential error accumulation caused by the sparse event data. Additionally, to address the problem of severely limited real dataset, we develop a new optical system to collect a real-world dataset with paired high-speed HDR videos and event streams, facilitating future research in this field. Our dataset provides the first real paired dataset for event-to-HDR reconstruction, avoiding potential inaccuracies from simulation strategies. Experimental results demonstrate that our method can generate high-quality, high-speed HDR videos. We further explore the potential of our work in cross-camera reconstruction and downstream computer vision tasks, including object detection, panoramic segmentation, optical flow estimation, and monocular depth estimation under HDR scenarios.

\end{abstract}

}

\maketitle

\IEEEdisplaynontitleabstractindextext

%
\IEEEpeerreviewmaketitle

\IEEEraisesectionheading{\section{Introduction}\label{sec:introduction}}

%
%
%
%
\IEEEPARstart{C}{ontrary} to conventional cameras that capture scene intensities at a fixed frame rate, event cameras employ a distinct approach by detecting pixel-wise intensity changes. A unique feature of event cameras is the asynchronous recording of events, triggered whenever a pixel's intensity change reaches a certain contrast threshold. Event cameras offer several advantages over traditional frame-based ones, such as low latency, low power consumption, high temporal resolution, and high dynamic range (HDR) \cite{Gallego2020tpami}. These features make event cameras beneficial for various vision tasks, including real-time object tracking \cite{ramesh2018long,zhang2021object,wang2024multi,liu2024siamese}, high-speed motion estimation \cite{lee2014real}, face recognition~\cite{lagorce2016hots}, optical flow estimation \cite{gallego2018unifying, shiba2022secrets,zhu2019unsupervised}, depth prediction \cite{zou2017robust, gehrig2021combining, mostafavi2021learning}, egomotion \cite{zhu2019unsupervised,ye2020unsupervised}, and on-board robotics \cite{waniek2015cooperative}.

Despite the numerous advantages offered by the unique triggering mechanism of event cameras, event data cannot be directly utilized in existing prevalent frame-based vision algorithms. The reason is that event cameras capture only changes in intensity, devoid of any absolute intensity values. Furthermore, the unique data format of event streams, represented as $4$-tuples, requires specialized processing pipelines that do not align with traditional image processing methodologies. This has led to a growing interest in converting event data into intensity images.

\begin{figure*}[h!] \small
	\centering
	\setlength{\tabcolsep}{1pt}
	\resizebox{0.98\linewidth}{!}{
	\begin{tabular}{cccccc}
		\includegraphics[width=.162\linewidth,clip,keepaspectratio]{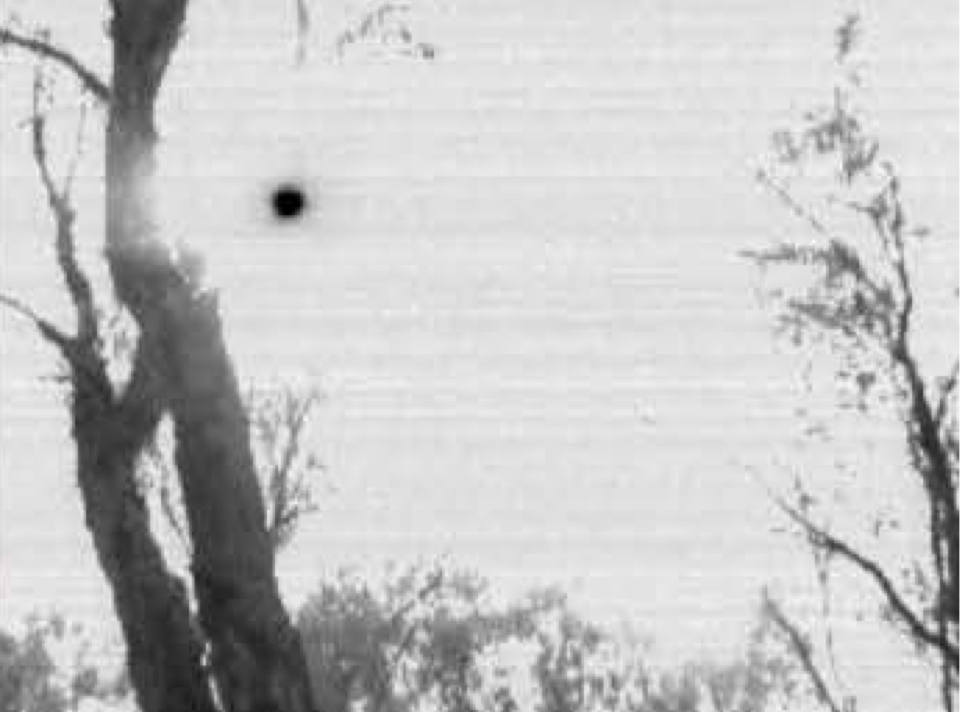} &
		\includegraphics[width=.16\linewidth,clip,keepaspectratio]{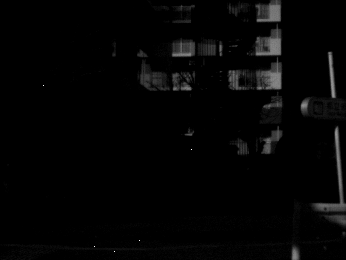} &
		\includegraphics[width=.16\linewidth,clip,keepaspectratio]{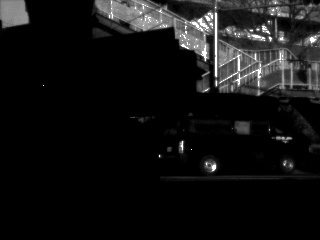} &
		\includegraphics[width=.16\linewidth,clip,keepaspectratio]{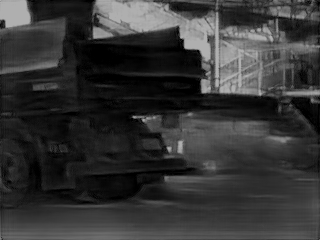} &
		\includegraphics[width=.16\linewidth,clip,keepaspectratio]{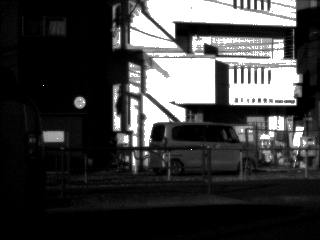} &
		\includegraphics[width=.16\linewidth,clip,keepaspectratio]{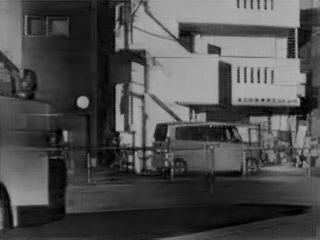} \\
		HDR Scene& HDR Scene& APS & Our Recon. & APS & Our Recon. \\
		\includegraphics[width=.162\linewidth,clip,keepaspectratio]{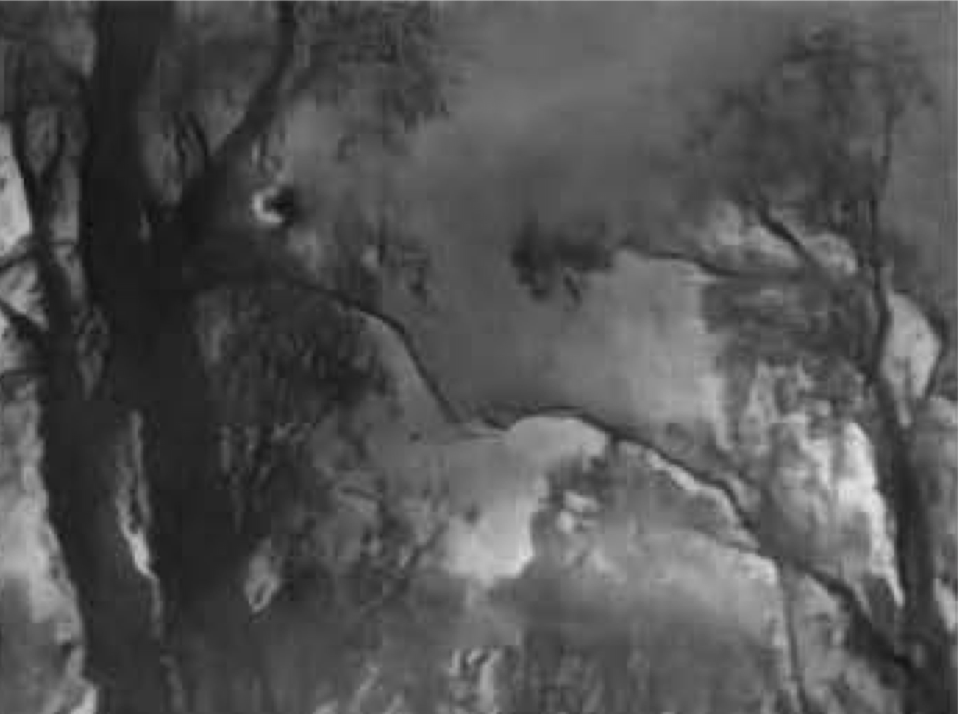} &
		\includegraphics[width=.16\linewidth,clip,keepaspectratio]{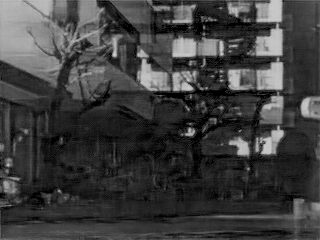} &
		\includegraphics[width=.16\linewidth,clip,keepaspectratio]{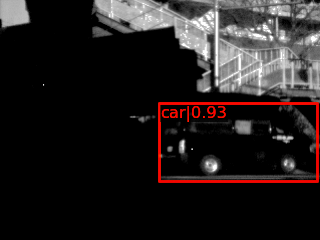} & 
		\includegraphics[width=.16\linewidth,clip,keepaspectratio]{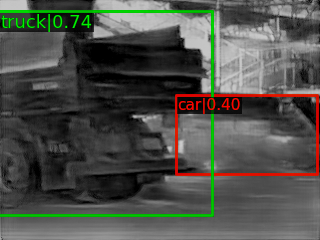} &
		\includegraphics[width=.16\linewidth,clip,keepaspectratio]{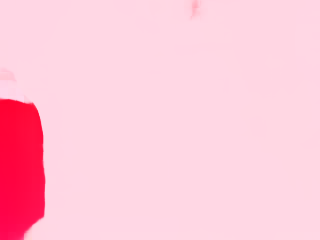} & 
		\includegraphics[width=.16\linewidth,clip,keepaspectratio]{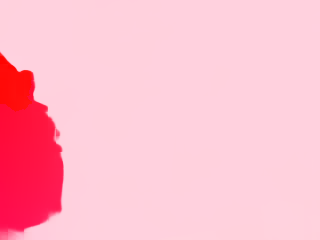}\\
		
		Pre. Recon.~\cite{rebecq2019high} & Our Recon. & APS detection & Our detection & APS flow & Our flow \\
	\end{tabular}}
	\vspace{-1mm}
	\caption{The comparison between previous  methods/APS and our method. Previous work~\cite{rebecq2019high} lacks real high bit-depth HDR images as ground truth, resulting in reconstructed images that suffer from severe artifacts. In contrast, APS images exhibit a low dynamic range, leading to suboptimal performance in downstream vision tasks under HDR scenes. Our method, however, produces visually pleasing results in both HDR reconstruction and real HDR applications, showing its superiority over existing approaches.}
	\label{fig:fig1}
	\vspace{-1mm}
\end{figure*}

Recent research efforts have concentrated on reconstructing high-speed~\cite{wang2019event, rebecq2019high} or HDR~\cite{wang2020event, weng2021event, yang2023learning,liu2023sensing} intensity images/videos from events, broadening the potential uses of event cameras. These methods employ deep neural networks to achieve state-of-the-art reconstruction performance. However, the commonly used recurrent processing approach for sequential data typically introduces error accumulation throughout the sequence. Additionally, they either neglect temporal constraints or employ a less accurate flow warping loss, both of which can negatively impact the video reconstruction quality. Consequently, the visual quality of the reconstructed high-speed HDR videos remains unsatisfactory. This inspires us to explore enhanced methods and systems for video reconstruction from events, and to better exploit the HDR capabilities of event data.

Another potential reason for the insufficient reconstruction quality could be the absence of high-quality training data. Notably, during training, existing researches \cite{wang2019event, rebecq2019high, wang2020event} consistently rely on simulating events using simulators like ESIM~\cite{rebecq2018esim} rather than using real event data. However, the motion of a virtual event camera may not accurately represent reality. Moreover, although significant efforts \cite{stoffregen2020reducing} have been made to improve the simulator, it remains unclear how well the simulated events correspond to real events captured by event cameras, particularly when considering complex factors such as noise~\cite{tian2023transformer,zhang2024deep} and data transfer bandwidth limitations present in real event cameras. Furthermore, the domain disparity between training and testing datasets precludes current methodologies from concurrently utilizing both the high-speed and HDR features of events. This motivates us to develop appropriate imaging devices that can capture paired high-speed HDR videos and events, thereby alleviating these constraints.

In this paper, we simultaneously exploit the dual virtues of event cameras, \ie, their exceptional frame rate and robust dynamic range tolerance, aiming to reconstruct high-speed HDR videos from event streams. To address the temporal sparsity in reconstructing such videos, we propose a recurrent neural network (RNN) guided by key frames, specifically designed for high-speed video reconstruction. Our innovative deep model is designed for extracting sequential information correlations along high-speed frames. To mitigate information loss in sparse event streams, we introduce a key frame guidance mechanism that feeds crucial data into the network. Additionally, we deploy a pyramidal deformable network to align features between consecutive event frames, with a temporal consistency constraint to enhance continuity across the sequence. 

Beyond the reconstruction network, we introduce a real paired dataset of \textit{event} and high-speed \textit{HDR} data, named EventHDR, utilizing our carefully designed imaging prototype. Importantly, we employ dual high-speed cameras and  HDR fusion to generate high-quality ground truth data, preserving the core \textit{high-speed} and \textit{HDR} characteristics of dynamic environments. The use of our real paired training data has greatly enhanced the reconstruction results in HDR scenes compared to existing methods~\cite{rebecq2019high}. As illustrated in Fig.~\ref{fig:fig1}, due to the unreal training data, the reconstructions of previous methods~\cite{rebecq2019high} suffer from severe artifacts and do not resemble real intensity images, while our model can generate visually impressive high-speed HDR videos from event streams. Fig.~\ref{fig:fig1} further exhibits a comparative analysis of our reconstruction and active pixel sensor (APS) performance in various HDR environments on downstream applications. Experimental results demonstrate that our method achieves state-of-the-art reconstruction performance, and incorporating paired real-world data in the training stage further assists the model in handling real HDR scenes for various computer vision tasks. 

A preliminary version of this work was presented as a conference paper~\cite{zou2021learning}. In this paper, the main extensions of our work can be summarized as follows
\begin{enumerate}
	\item We develop a recurrent convolutional network tailored for the high-precision reconstruction of high-speed HDR videos from event data, incorporating  a novel key frame guidance mechanism to mitigate information loss and a local attention fusion module to efficiently handle the temporal-spatial correlations in highly sparse and high-speed event data.
	\item With our innovative co-axis imaging system, we enhance EventHDR dataset by improving quality, quantity, and the diversity of scenarios. This new dataset gathers spatially and temporally aligned high-speed HDR videos along with corresponding event streams, providing a unique data preparation that supersedes traditional numerical simulation. Extensive discussions have confirmed the dataset's indispensability for the Event-to-HDR task.
	\item Our approach opens the door for practical applications of events in various HDR scenarios. The capability of our model has been tested through tasks such as object detection,  panoramic segmentation, optical flow estimation and monocular depth estimation, providing a comprehensive analysis of the potential uses of our model. Our in-depth exploration of Event-to-HDR network design presents yields insights to future research in this domain.

\end{enumerate}

\vspace{-2mm}
\section{Related Work}
This section provides an overview of research works closely related to this paper. We begin with  explorations of intensity image and video reconstruction techniques. Subsequently, we delve into applications of HDR using events.
\vspace{1mm}

\noindent\textbf{Intensity Images and Videos Reconstruction from Events.}
While event-based cameras offer numerous advantages over traditional imaging methods, such as higher temporal resolution and dynamic range, their practicality is limited in many applications. Specifically, the asynchrony and stream nature of event data hinder their direct use in a range of existing computer vision algorithms. Recognizing the potential to harness both the distinctive features of event cameras and the power of contemporary computer vision techniques, there has been a growing interest in reconstructing intensity images and videos from event data.

In early research on event-to-image reconstruction, Cook \etal \cite{cook2011interacting} proposed a network that used recurrently interconnected areas to interpret events, reconstructing light intensity and optical flow. Kim \etal \cite{kim2014simultaneous} created a high-resolution mosaic using probabilistic filtering. Recently, deep learning has been applied to event-based image and video reconstruction, achieving significant results. To meet the data prerequisite for deep models, many approaches generate synthetic event training data via the ESIM simulator~\cite{rebecq2018esim}, with input images or APS images as ground truth. Examples include that Rebecq \etal \cite{rebecq2019high} employed a convolutional recurrent neural network with flow warping loss, and Wang \etal \cite{wang2019event} presented a generative adversarial network. Other works, such as EventSR~\cite{wang2020eventsr}, aimed at reconstructing, restoring, and super-resolving images simultaneously. Mohammad \etal~\cite{I._2020_CVPR} focused on super-resolving event data to high-resolution intensity images. In the recent past, advanced deep learning architectures have been proposed~\cite{ye2022unsupervised,li2024supervise} . Weng \etal \cite{weng2021event} proposed a hybrid CNN-Transformer network called ET-Net, harnessing both the local and global contexts inherent in event sequences. Sabater \etal\cite{sabater2022event, sabater2023event} also utilized transformer architecture, crafting a solution that is both light-weight and sparsely designed. On a different way, Zhu \etal~\cite{zhu2022event} first introduced deep spiking neural networks for computationally efficient video reconstruction from events. While these  methods effectively build images from events, visual quality, especially in HDR scenes, remains limited due to the lack of real HDR data during training. Despite efforts to improve synthetic data quality~\cite{stoffregen2020reducing,luo2023learning} or event simulators~\cite{lin2022dvs, li2023blinkflow}, the absence of paired real HDR training data remains a challenge.

\vspace{1mm}

\noindent\textbf{HDR Reconstruction from Events.}
Event cameras record intensity changes on a logarithmic scale, granting them heightened sensitivity in exceptionally dark scenarios while remaining resistant to intensity overflows. As a result, event streams are less affected by overexposure in bright conditions compared to ordinary consumer cameras~\cite{fu2023raw}, making them as particularly adept for HDR scenes. Taking this advantage, Kim \etal~\cite{kim2014simultaneous} pioneered HDR image reconstruction from event streams by creating a high-resolution, high dynamic range mosaic of a scene under the assumption of rotational camera motion. Subsequently, many learning-based event-to-image reconstruction methods~\cite{wang2019event, rebecq2019high, I._2020_CVPR} trained their models on ordinary images and directly generalized their models to HDR scenes during the testing stage. Although their reconstruction results reveal details in dark and bright regions, the visual perception of these reconstructions often diverged from real scenes. This phenomenon is caused by their experimental settings, which use only LDR training samples to simulate events. To address this issue, Han \etal~\cite{han2020neuromorphic} proposed a hybrid HDR imaging system that fuses an LDR image with an intensity map obtained from the corresponding event streams to create an HDR image. Their results appear more visually natural, but the low capture speed of LDR frames restricts its application to more high-speed case. In summation, prevailing techniques grapple with seamlessly integrating both the high-speed and HDR facets inherent to event streams.

\vspace{1mm}
\noindent\textbf{Downstream Vision Applications Using Events.}
The potential of event streams has been harnessed by researchers to address an array of computer vision challenges, such as real-time object tracking \cite{ramesh2018long, zhang2021object,wang2024multi}, high-speed motion estimation \cite{lee2014real}, optical flow estimation \cite{gallego2018unifying, shiba2022secrets, zhu2019unsupervised}, depth estimation \cite{zou2017robust, gehrig2021combining, mostafavi2021learning}, egomotion estimation \cite{zhu2019unsupervised, ye2020unsupervised}, and on-board robotics \cite{waniek2015cooperative}. Nonetheless, the direct incorporation of event data into such tasks can be inconvenient, as each task needs specially designed and  trained models directly on event data, which limits the mobility and flexibility. An alternative approach involves converting events to intensity images first and then applying well-established frame-based algorithms on them~\cite{mostafavi2021learning}. This methodological shift simplifies the process: by focusing primarily on the reconstruction task, subsequent tasks can be addressed using existing solutions. However, this sequence places significant weight on the quality of HDR reconstruction. The effective harnessing of the high-speed and HDR characteristics of event streams becomes crucial, with profound implications for the outcomes of subsequent tasks.

In our research, we circumvent the issues tied to the inconvenient application of event data in vision tasks. Through our innovative reconstruction network and real paired training dataset, we enhance HDR video reconstruction quality, thereby boosting the performance of downstream vision tasks.

\section{Event-to-HDR Architecture Design}
\label{sec:method}
In this section, we first define our problem and outline the motivation behind our approach. We then detail the strategy for representing events. Next, we introduce our model architecture and conclude with specific implementation details.

\begin{figure*}
   \centering
	\includegraphics[width=0.9\linewidth, clip, keepaspectratio
	]{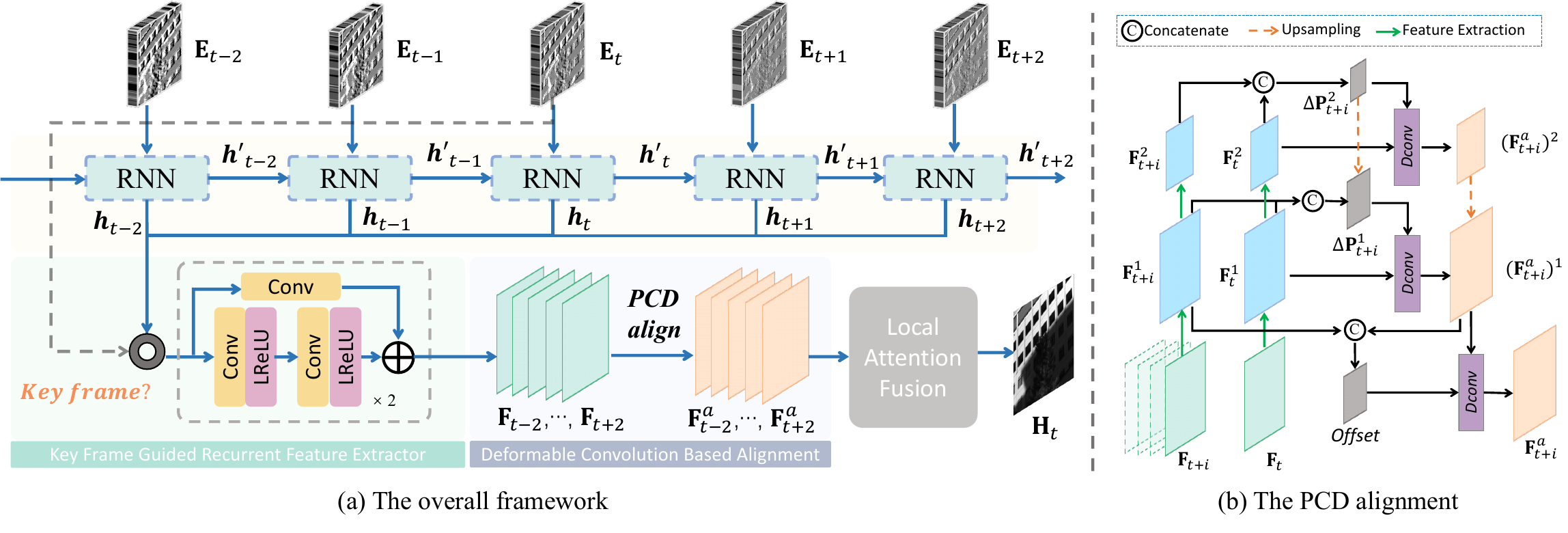}
	\vspace{-5mm}
	\caption{The overview of our recurrent convolutional neural network for HDR video reconstruction from events.}
	\label{fig:overview}
\end{figure*}
	\vspace{-2mm}

\subsection{Formulation and Motivation}
Event cameras capture a continuous stream of asynchronous spikes. An event is triggered when the logarithm of the brightness change at a specific pixel surpasses a predefined contrast threshold. As a result, each event data is captured as a $4$-tuple tensor
\begin{equation}
	\bm e=(x, y, q, t),
\end{equation}
where $(x, y)$ are the pixel coordinates, $q\in{\pm 1}$ represents the polarity, and $t$ indicates the timestamp. Let $S$ symbolize the contrast threshold and  $I_{xy}(t)$ be the intensity at time $t$ for a pixel at location $(x, y)$. The process of event generation can then be depicted as
\begin{equation}
	\log(I_{xy}(t)) - \log(I_{xy}(t-\Delta t)) = qS,
\end{equation}
where $t-\Delta t$ is the timestamp of last event at location $(x, y)$.  The captured data of an event camera is a set of continuous event streams 
$\{\bm e_i\}$.

Given that event cameras capture intensity changes on a logarithmic scale, they are specially good at recording details in low-light conditions~\cite{chen2023instance} while being less prone to overexposure. Consequently, event cameras are particularly well-suited for handling HDR scenes. Although previous works~\cite{wang2019event, rebecq2019high, I._2020_CVPR} have shed light on the potential of events in HDR image and video reconstruction, they primarily aim at managing HDR scenes rather than generating standard high-bit HDR images. Their focus on differential information over absolute scene intensity leads to the risk of artifacts, potentially reducing the realism of the reconstructed outputs. In addition, the extremely long and sparse event sequences pose more difficulties for existing methods due to information loss. Thus, how to extract the sparse information is of vital importance. To circumvent these limitations, we present a novel approach specifically tailored for high-speed HDR (high-bit) video reconstruction. Given an event stream $\mathcal {E}=\{\bm e_i\}$ and the corresponding ground truth HDR frames $\mathcal{H}=\{\mathbf H_t\}$, our goal is to reconstruct high-speed HDR video $\mathcal{H}$ from the event stream $\mathcal {E}$ using an end-to-end neural network architecture. We present a recurrent neural network, with careful alignment to fully extract information from sequential data. Moreover, we incorporate a key-frame guiding approach to counteract the potential data deficiencies and error accumulation in long, sparse event sequences.

\vspace{-1mm}
\subsection{Event Stacking}
Convolutional neural network is a powerful tool to deal with images and sequential videos. A popular approach for dealing with event data is to embed event streams into voxel grids, also referred to as event frames that have spatial features similar to conventional image frames. Numerous methodologies for integrating event streams into tensors have been exploited in the past, including temporal-based stacking, event-number-based stacking, and inter-frame stacking~\cite{wang2019event, rebecq2019high, stoffregen2020reducing}. In video reconstruction, it is natural to stack events between two reference frames. This ensures a uniform timestamp across the reconstructed frames. In order to make use of the well developed deep convolutional network architecture, transforming event streams into 3D spatial volumes is required. Previous works \cite{rebecq2019high, stoffregen2020reducing, wang2019event} have presented several ways to stack event frames into spatio-temporal tensor. Gehrig~\etal~\cite{gehrig2019end} divided those grid-based representations into $6$ categories, including event frame, event count image, surface of active events, voxel grid, histogram of time surfaces and event spike tensor. The study of \cite{rebecq2019high} have proved the superiority of event spike tensor, and in this work, we follow their work to use event spike tensor representation strategy. Specifically, denoting the duration between two consecutive ground truth frames as $\Delta T$, the events between are firstly divided into $B$ temporal bins. Then, the polarity of each event is separated into the  two neighboring bins, which can be expressed as
\begin{equation}
\bold E_{\pm}(x_l, y_m, t_n) = \sum_{{e_i \in \mathcal E_{\pm}}} \max(0, 1 - \vert t_n - t_i^*\vert),
\end{equation}
\begin{equation}
 t_i^* = \frac{B-1}{\Delta T}(t_i - t_0),
\end{equation}
where $(x_l, y_m, t_n)$ is the coordinates that cover the entire spatio-temporal scope. $t_n\in\{t_0, t_0+\Delta T, \cdots, t_0+(B-1)\Delta T\}$, $\mathcal E_+$ and $\mathcal E_-$ represent the positive and negative events within the duration $\Delta T$, and $t_i^*$ denotes the normalized event timestamp for event $e_i$. In this way, for each polarity, we obtain a $B$-channel tensor, thus the asynchronous event streams are represented as grid-like synchronous tensors with $2B$ channels which contains spatial information.

\subsection{Network Architecture}
\label{sec:network}
To address the challenges posed by high-speed sparse event frames, it is essential to fully harness the sparse information across long sequences. In this study,
we introduce a recurrent convolutional neural network, guided by key frames, to reconstruct HDR videos from an event stream, as illustrated in Fig.~\ref{fig:overview}. Our model processes $T=2N+1$ consecutive event voxel grids $\{\mathbf E_{t-N}, \hdots, \mathbf E_{t+N}\}$ to construct the HDR frame $\mathbf H_t$ at timestamp $t$. The network initially employs a key frame guided recurrent feature extractor to obtain sequential features from various event frames. Subsequently, these features are aligned via a deformable convolution-based module and are then fused through a local attention block. Additionally, we introduce a novel consistency loss mechanism to maintain temporal continuity. 

\subsubsection{Key Frame Guided Recurrent Feature Extractor} The event-to-HDR reconstruction problem is highly ill-posed, given that events only capture differential information about the scene and lack absolute intensity values. This issue is exacerbated when reconstructing extremely high frame rate videos, as the information between two consecutive ground truth frames is minimal. Independently extracting features at different timestamps could lead to insufficient spatial information for the network. To address this, we employ a recurrent feature extractor designed to exploit sequential information over a more extensive temporal range. In this module, we use a recurrent neural network to propagate temporal information to features at different timestamps. The key frame guided recurrent feature extractor can be formulated as
  \begin{equation}
  	\{\mathbf F_t', \bm h_i \}= W(\bold E_i, \bold E_{i-1}, \bm h_{i-1}),
  	\label{eq:recurrent}
  \end{equation}
where $\mathbf F_t'$ is the extracted feature, $\bm h_i$ denotes the hidden state and $W$ represents the feature extractor. In our work, $W$ is composed of several strided convolutional networks to downsample  the original input tensor to a lower spatial resolution. In this way, the computation cost is reduced while remaining the most useful information of the events. 

The recurrent architectural design, though advantageous in sequence handling, is susceptible to error propagation as sequence length increases. Therefore, we propose a key frame guidance strategy to address this issue. Typically, we designate certain frame indices $\mathcal K$ as \textit{key frames}. In these frames, the extracted features undergo a refreshment process by integrating input event frame, which aids in maintaining the continuity of the image sequence while also adding timely corrections that help stave off accumulated errors. The key frame guidance can be formulated as
\begin{equation}
	\mathbf F_t=\left\{
	\begin{aligned}
		G(\mathbf F_t', \mathbf E_t)  \quad \quad\quad \quad t \in\mathcal K\\
		\mathbf F_t'   \quad \quad\quad\text{otherwise},
	\end{aligned}
\right.
\end{equation}
where $G$ is residual blocks. In the experiment, we set $\mathcal K$ to a multiple of 5, \ie, $\{0, 5, 10, ...\}$.

\subsubsection{Deformable Convolution Based Feature Alignment} 
Conventional approaches to event-to-HDR reconstruction typically rely on optical flow~\cite{I._2020_CVPR} to align disparate frames or employ a flow-warping loss~\cite{rebecq2019high} to mitigate temporal discrepancies.. However, accurately extracting flow in scenarios where event streams are sparse and differ significantly from common intensity images poses a considerable challenge, often leading to motion anomalies~\cite{tassano2020fastdvdnet}. More severely, given the sparse nature and distinct format of event streams compared to conventional intensity images, achieving precise flow predictions becomes even more. Therefore, a more robust alignment approach with better learnability are needed to deal with event data. 

To address these limitations and inspired by relevant works in video restoration and object detection~\cite{wang2019edvr, dai2017deformable}, we have integrated pyramidal deformable convolutions into our alignment strategy. This approach enhances the adaptability of traditional convolutional kernels by optimizing their offsets, thereby improving feature alignment across frames. The choice to employ pyramidal deformable convolutions is driven by their proven effectiveness in addressing misalignments by learning offsets directly through the network, allowing for adjustments tailored to the dynamic specifics of each scene~\cite{wang2019edvr}. Beyond calculating convolution offsets within neighboring windows~\cite{wang2019edvr}, our model incorporates long-range sequence information into the offset computation. This capability enables the alignment module not only to leverage information from adjacent frames but also to utilize dependencies over longer sequences. By incorporating these broader temporal relationships, our approach significantly enhances the accuracy and robustness of the alignment process, making it ideally suited for the complexities of event-to-HDR video reconstruction.

In the alignment module, our goal is to align features of different event frames $\mathbf F_{t+i}$ to the feature of the central frame $\mathbf F_t$. Assuming that a convolution kernel has $K$ locations. Take a common $3\times 3$ kernel for example, we have $K=9$ and the regular grid $\mathcal R=\{(-1,-1), (-1, 0), \hdots, (1,0), (1,1)\}$, which denotes the locations of an ordinary convolution operation. For each location $\mathbf p_0$ on the output feature map, the aligned feature can be expressed as
\begin{equation}
	\mathbf F_{t+i}^{a}(\mathbf p_0) = \sum_{\mathbf p_j \in \mathcal R}\mathbf w(\mathbf p_0)\cdot \mathbf F_{t+i}(\mathbf p_0+\mathbf p_j+\Delta \mathbf p_j),
	\label{eq:dconv}
\end{equation}
where $\mathbf w$ is the weights for each location in $\mathcal R$, and $\mathbf p_j$ and $\Delta \mathbf p_j$ denote the pre-specified offset and learnable offset of $j$-th location in deformable convolutions. Eq.~\eqref{eq:dconv} illustrates the operation of a simple deformable convolutional layers that the convolutions are sampled on an extra offset $\Delta \mathbf p_j$, comparing to ordinary convolutional layers.

To predict the learnable offset $\Delta \mathbf P=\{\Delta \mathbf p_j\}_{\mathbf p_j \in \mathcal R}$ for the $(t+i)$-th event feature, the feature of the $(t+i)$-th frame $\mathbf F_{t+i}$ and the central frame $\mathbf F_{t}$ are sent to the offset predicting operation $f$, and can be expressed as
\begin{equation}
	\Delta \mathbf P_{t+i}=f( \mathbf F_{t+i}, \mathbf F_{t}).
\end{equation}
we employ pyramidal processing and cascading refinement to enlarge the receptive field of the offsets and align larger movement like \cite{wang2019edvr, yue2020supervised}. Specifically, assuming that the pyramidal architecture consists of $L$ levels, and the feature of the $l$-th layer $\mathbf F_{t+i}^{l}$ is downsampled through strided convolutions with a factor of 2 on the $(l-1)$-th feature $\mathbf F_{t+i}^{l-1}$. After obtaining all of the $L$ features, we calculate the offset for the $l$-th layer from the upsampled $(l+1)$-th offsets and the $l$-th pyramidal feature, as shown in Fig.~\ref{fig:overview}. This process can be interpreted as
\begin{equation}
	\Delta \mathbf P_{t+i}^{l} = f(\mathbf F_{t+i}, \mathbf F_{t}, \mathcal U(\Delta \mathbf P_{t+i}^{l+1})),
\end{equation}
where $\mathcal U$ denotes bilinear upsampling operation. Thus, the aligned feature  of the $l$-th level can be expressed as 
\begin{equation}
	(\mathbf F_{t+i}^{a})^l=g(\text{DConv}(\mathbf F_{t+i}^l, \Delta \mathbf P_{t+i}^{l}), \mathcal U((\mathbf F_{t+i}^{a})^{l+1})).
	\label{eq:alignedfeature}
\end{equation}
In Eq.~\eqref{eq:alignedfeature}, $g$ is convolutional layers to generate aligned features, and $\text{DConv}$ is the deformable convolution described in Eq.~\eqref{eq:dconv}. In this way, we obtain the aligned feature $(F_{t+i}^{a})^1$ for the $1$-st pyramidal layer. We further use the feature $F_t^1$ of the reference frame to generate the final aligned feature $F_{t+i}^{a}$ from $(F_{t+i}^{a})^1$. For each of the $T$ frames, we could obtain the corresponding aligned feature from  Fig.~\ref{fig:overview}.

\subsubsection{Local Attention Based Feature Fusion} After obtaining the aligned event features for consecutive $2N+1$ frames, we need to further leverage these features to form a unified feature for final reconstruction.  Along the temporal sequence, different features contains varied information, and our goal is to aggregate neighboring features. Attention mechanism can be used to stress the significance of different event frames or spatial  locations, and in this work, we introduce a key matching based local attention mechanism to fuse temporal features. Specifically, we construct a key $ K_i \in R \times C \times D$ for each feature $\bold F_i^a$, and fuses these features by matching keys. Here we use a simple dot product to measure the correspondence between keys. When reconstructing the video frame at timestamp $t$, we match the keys $ K_i $ from all neighboring frames with the central frame $ K_t$, and obtain a 4D attention map $A_{it}(m, n, u, v)$ which records the similarity between pixel $(m, n)$ of $\bold F_i^a$ and pixel $(u, v)$ of $\bold F_t^a$, as expressed as
  \begin{equation}
  	 A_{it}(m, n, u, v) =  K_i(m, n)^\text T  K_t(u, v).
  	 \label{eq:local_attention_map}
  \end{equation}
Considering that reconstructing extremely high-speed videos is computation consuming, we only calculate a local attention map with a radius of $L$. Therefore, in Eq.~\eqref{eq:local_attention_map}, $(m, n) \in \{1, \cdots, R\}\times\{1, \cdots, C\}$, $(u, v) \in \{R-L, \cdots, R+L\}\times\{C-L, \cdots, C+L\}$. Since $L<<R, C$, the number of network operations are reduced largely.

The attention map $A_{it}(m, n, u, v)$ contains the correlation between frame $i$ and the reference frame $t$, we transform the attention map  into a normalized similarity matrix, which can be represented as
\begin{equation}
	P_{it}(m, n, u, v) = \frac{\exp( A_{it}(m, n, u, v))}{\sum_{ipq}\exp(A_{it}(m, n, u, v))}.
\end{equation}
Then, with the neighboring feature $\bold F_i^a$ and the  similarity matrix $P_{it}$ which infers probability, the feature for reconstruction can be obtained by a weighted summation of all neighboring features as
\begin{equation}
	\widetilde{\bold F}_t(m, n) = \sum_i \sum_{uv} P_{it}(m, n, u, v)\bold F_i^a(m, n).
\end{equation}

\begin{figure*}[ht]
	\centering
	\subfloat[Optical system]{
		\includegraphics[width=0.32\linewidth]{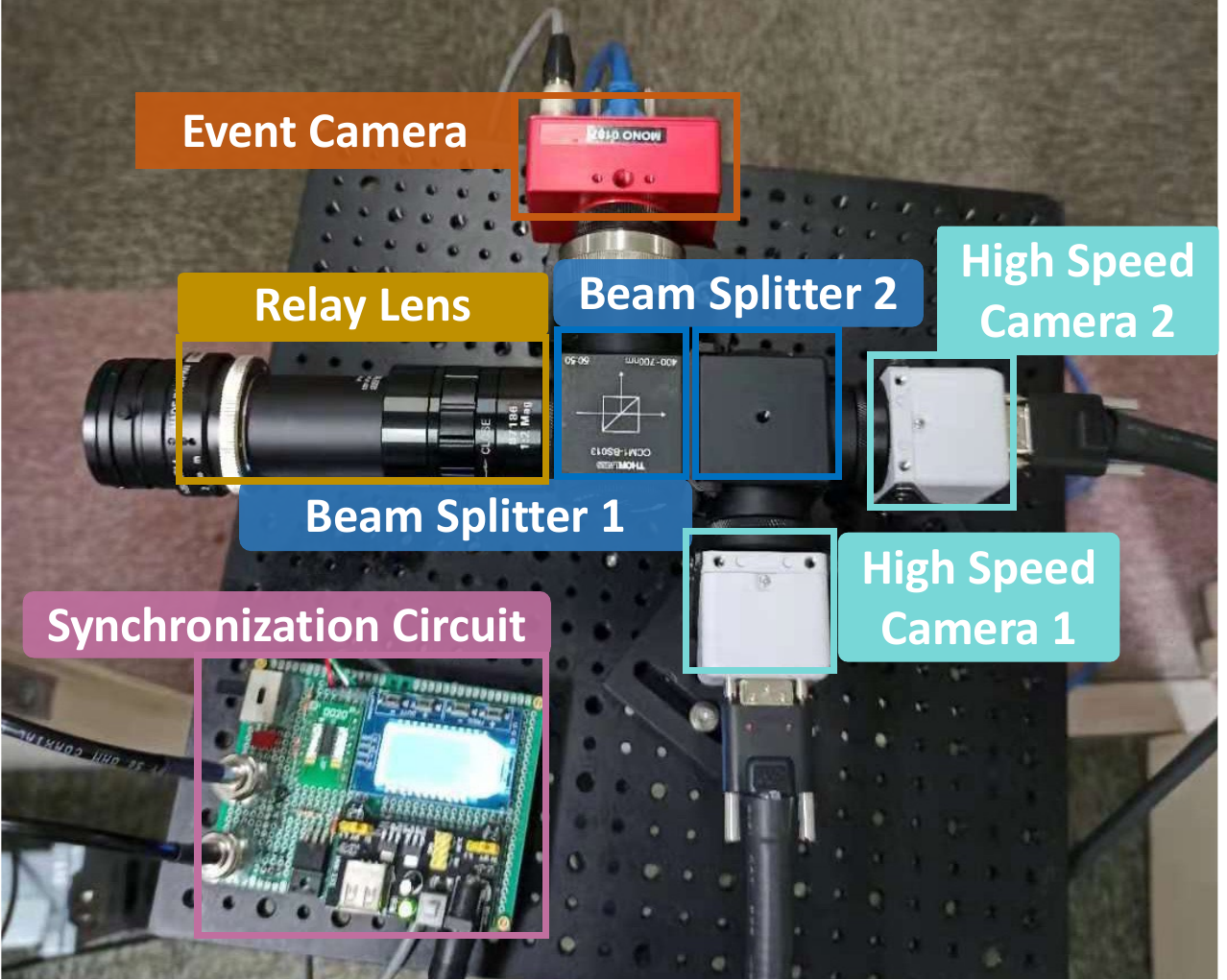}
	}
	\subfloat[Electronic system]{
		\includegraphics[width=0.33\linewidth]{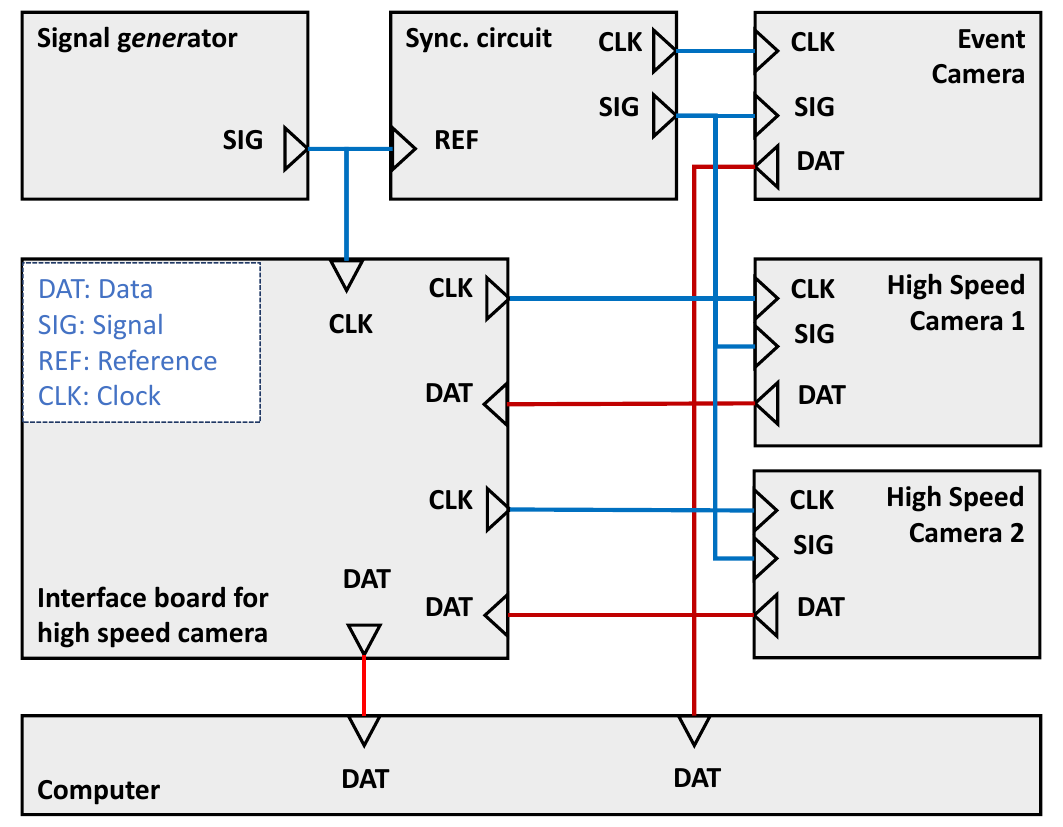}
	}
	\subfloat[Data synchronization]{
		\includegraphics[width=0.225\linewidth]{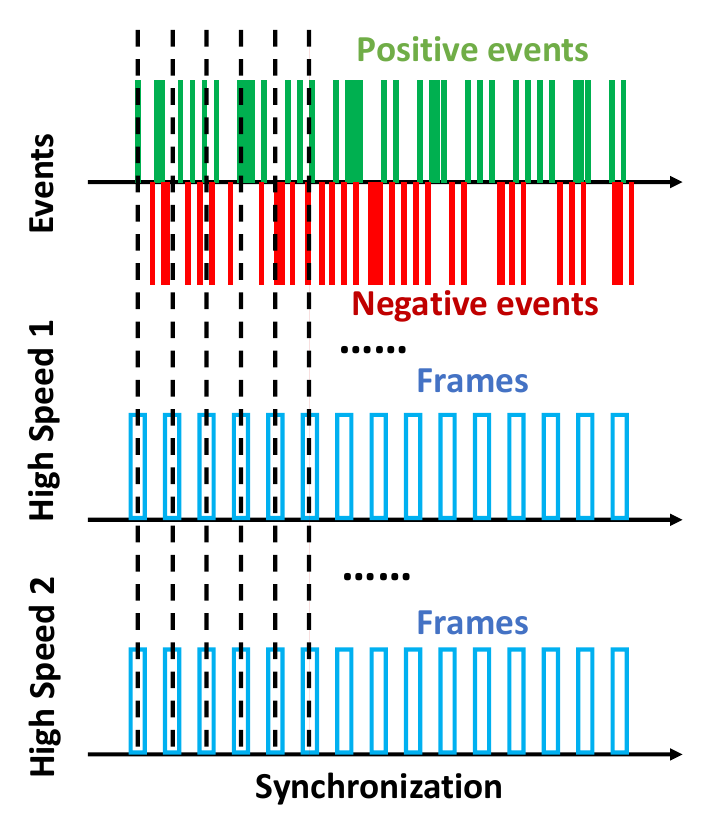}
	}
\vspace{-1mm}
	\caption{The hardware implementation of our high-speed HDR video imaging system with events. (a) The optical system. It contains two high-speed cameras and an event camera. (b) The electronic system. It controls three cameras to synchronously capture the information of same scene. (c) The synchronize data captured by three cameras, respectively.}
	\label{fig:hardware}
\end{figure*}
	\vspace{-3mm}

\vspace{-2mm}
\subsection{Learning Details}
Previous works~\cite{rebecq2019high, I._2020_CVPR} employed flow warping error~\cite{lai2018learning} for temporal consistency loss. These approaches, however, can sometimes yield issues stemming from less accurate flow estimations. In light of these challenges, we introduce a novel approach to compute temporal consistency loss. 
Several works~\cite{pan2019bringing, wang2020event,pan2020high} have analyzed that the intensity change of two successive sharp frames can be represented by the integral of events between these two frames.  Building upon this understanding, we derive a temporal consistency loss formula that emphasizes maintaining temporal continuity along reconstructed sequences. For two successive ground truth frames, $\hat {\mathbf H}_{t-1}$ and $\hat {\mathbf H}_{t}$, the corresponding event frame $\mathbf E_t$ can be inferred as
\begin{equation}
	\mathbf E_t = \mathcal C(\hat {\mathbf H}_{t-1}, \hat {\mathbf H}_{t}),
\end{equation}
Here, $\mathcal C$ represents the integral relationship bridging frames with their events. Intuitively, we could simply regard $\mathcal C$  as a process similar to the ESIM simulator~\cite{rebecq2018esim}, but a precise derivation from training data provides enhanced accuracy.
To map frames to events, we use the UNet-like convolutional neural network \cite{unet2015}. Before the primary model training phase, this network is pre-trained, functioning as a temporal consistency loss module to ensure that consecutive reconstructions closely mirror actual scenes. The temporal consistency loss can be expressed as
\begin{equation}
	\mathcal L_{C} = \sum_{i=1}^{T}\Vert \mathbf E_{t} - \mathcal C(\mathbf H_{t-1}, \mathbf H_{t}) \Vert_2^2.
\end{equation}
For a reconstructed video sequence, $\mathbf H_i$, and its ground truth, $\hat {\mathbf H}_i$, we utilize the standard $l_1$ loss for assessing reconstruction
\begin{equation}
	\mathcal L_{l_1} = \sum_{i=1}^{T}\Vert \mathbf H_i-\hat {\mathbf H}_i\Vert.
\end{equation}
Considering that relying solely on the $l_1$ reconstruction loss could introduce blurry distortions, we supplement it with the Learned Perceptual Image Patch Similarity (LPIPS~\cite{zhang2018unreasonable}) loss to ensure higher-level and structural similarity. Thus, the comprehensive loss function for high-speed HDR reconstruction from events becomes
\begin{equation}
	\mathcal L=\mathcal L_{l_1} + \tau_1\mathcal L_{LPIPS} + \tau_2\mathcal L_{C}.
\end{equation}

 In our deformable convolution-based feature alignment module, through empirical hyper-parameter testing, we select a pyramid level of $L=3$. This choice strikes a balance between computational efficiency and performance improvement. During experiments, we set $\tau_1$ and $\tau_2$ to $2$ and $0.2$, respectively. Losses are optimized via the adaptive moment estimation method~\cite{kingma2014adam}, setting the momentum parameter at 0.9. The initial learning rate is $10^{-4}$ and undergoes a tenfold reduction every 50 epochs. With a batch size of 4, the training extends over 100 epochs. The model is crafted using the PyTorch deep learning framework~\cite{paszke2019pytorch}, and NVIDIA RTX 3090 GPUs power our training process.

\section{Imaging System and Dataset}
In this section, we introduce  the construction of our real-world HDR video and events imaging system, and provide the details of our EventHDR real-world dataset for Event-to-HDR  reconstruction.
\subsection{A Real-World HDR Video \& Events Imaging System}

Here, we build a novel imaging system for paired high-speed HDR videos and the corresponding event streams. Despite the potential value of real paired data, it remains unexplored in existing research due to the following reasons. Primarily, the act of recording high-speed HDR videos is not trivial. To meet the high-speed requirement, specialized high-speed cameras are needed. In addition, HDR generation in such high-speed condition is even more difficult. Synchronizing the timestamps and field of view between high-speed HDR and event cameras also poses problems, given that both cameras capture different modal information and the susceptibilities to noise~\cite{zou2022estimating, zou2023iterative,lai2024hyperspectral}.

To address these challenges, we design an elaborate system to synchronously capture paired high-speed HDR video and the corresponding event stream. In simple terms, our novel imaging system is a three-camera co-axis system. We employ an event camera to capture event streams and use two high-speed cameras to capture synchronized LDR frames, which are later merged to form an HDR frame. All three cameras share the same light path and field-of-view, ensuring that the same scene is captured. By carefully aligning these cameras through our meticulously designed system, we can record paired high-speed HDR videos and corresponding event streams. Our entire hardware prototype is illustrated in Fig.~\ref{fig:hardware}. As shown in Fig.~\ref{fig:hardware}(a), light from the scene first travels through a relay lens. We then use a Thorlabs CCM1-BS013 beam splitter to divide the incident light into two equivalent components with different directions. For one direction, an iniVation DAVIS346~\cite{Neuromorphic} event camera captures the event stream. For the other direction, another beam splitter is employed to further transmit the light to two Photron IDP-Express R2000 high-speed cameras, which capture two synchronous videos.

Following the HDR generation methodology presented by \cite{debevec2008recovering}, which synthesizes multiple LDR images at varied exposure times, we equip one of our high-speed cameras with a Thorlabs ND513B neutral density (ND) filter to reduce the incoming irradiance. An ND filter attenuates the incoming light across both spatial and spectral dimensions, thereby diminishing the scene's irradiance. Such an arrangement allows us to obtain two LDR images exhibiting different scene irradiance levels without manipulating exposure times, which often proves challenging in high-speed video captures. In our setup, the ND filter is chosen for its capability to attenuate approximately $90\%$ of the scene's irradiance. Given that both our high-speed cameras produce 12-bit images, the merging process yields an enhanced 16-bit HDR image. For these three cameras, the fields of view are strictly aligned for spatial accuracy. In addition, to guarantee the temporal synchronization of the three cameras, the timestamps are controlled by a specially designed circuit, as shown in Fig.~\ref{fig:hardware}(b)(c). Specifically, the capture of the three cameras is controlled by the electronic circuits. When the circuits deliver a triggering signal, the two high-speed cameras are triggered to capture the scene immediately. Meanwhile, the exact timestamp of the trigger is recorded in the metadata of the event stream. 

\subsection{The Proposed EventHDR Dataset} 
\label{sec:real_data}
An ideal dataset for high-speed HDR video reconstruction should meet several criteria, i.e., high-speed, high bit-depth~\cite{zou2023rawhdr}, HDR, and dynamic in both background and foreground. However, contemporary HDR video datasets~\cite{FroehlichSPIE2014, kronander2013unified, song2016sjtu} do not fulfill all these requirements simultaneously. To enhance the performance of our model when reconstructing high-speed HDR videos, we harness our imaging system to capture a high-quality, real video HDR dataset, dubbed EventHDR, for both training and evalutaion.

Through great efforts in imaging system design and data capture, our EventHDR dataset consists of  $26$ typical outdoor scenes for training. The scenes exhibit a high dynamic range, containing extreme illumination regions that cannot be accurately recorded by conventional cameras due to overexposure or loss of details in dark/bright areas. Each video lasts $5.6$ seconds with $2828$ frames, indicating an acquisition speed of $500$ fps. Cumulatively, our synchronized real dataset contains over $70,000$ HDR video frames, offering a robust foundation for training event-to-HDR networks.

Moreover, we gather $19$ videos for evaluation. Comprising both the original event streams and the ground-truth high bit-depth HDR sequences, this evaluation set promises to probe into the far reaches of extreme HDR imaging, while also facilitates other high-level tasks in HDR scenes. We offer a preview of our paired real-world dataset in Fig.~\ref{fig:real_data}, which presents the three co-axis cameras, as well as the input and output data for our EventHDR dataset. Further details on our imaging system and the EventHDR dataset can be accessed in Table~\ref{tab:data}. Additionally, as shown in Table~\ref{tab:datacompare}, EventHDR is the first real high bit-depth paired dataset with the highest frame rate to fully utilize event camera's high-speed feature, compared with existing Event-to-HDR reconstruction datasets~IJRR\cite{stoffregen2020reducing}, HQF~\cite{mueggler2017event}, and other event-based tasks tasks BS-ERGB~\cite{tulyakov2022time} and PIR2000~\cite{ding2022event}.

\begin{table}[t]
	\centering
	\setlength{\tabcolsep}{3mm}
	\footnotesize
	\caption{The details of our imaging system, and our real EventHDR dataset.}\label{tab:data}
	\vspace{-2mm}
	\begin{tabular}{|p{4.2cm}<{\centering}|p{3.5cm}<{\centering}|}
		\hline
		 Event  Camera  & iniVation DAVIS346        \\ \hline
		 Intensity Camera & Photron IDP-Express R2000 \\ \hline
		 Beam Splitter & Thorlabs CCM1-BS013 \\ \hline
		 ND filter & Thorlabs ND513B (90\%) \\ \hline
		 Original LDR Bit-Depth       & 12bit                      \\ \hline
		 Fused HDR Bit-Depth      & 16it                     \\ \hline
		 Frame Rate   &  500-2000 fps \\ \hline
		 Training Size & 26 sequences \\ \hline
		 Training Frames/Sequence      & 2828  \\ \hline
		 Training Sequence Length   & 5.6 seconds \\ \hline
		 Evaluation Size & 19 sequences \\ \hline  
		 Evaluation Frames/Sequence      & 400  \\ \hline
		 Evaluation Sequence Length   & 0.8 seconds \\ \hline
	\end{tabular}
\vspace{-2mm}
\end{table}

\begin{table*}[h!]
	\renewcommand{\arraystretch}{1.}
	\centering
	\footnotesize
	\setlength{\tabcolsep}{6mm}
	\begin{threeparttable}
		\caption{The comparisons of our real EventHDR dataset compared with other existing dataset in event-based vision.}
		\vspace{-2mm}
		\label{tab:datacompare}
		\begin{tabular}{cccccc}
			\hline\hline
			& IJRR~\cite{mueggler2017event} & HQF~\cite{stoffregen2020reducing} & BS-ERGB~\cite{tulyakov2022time} & PIR2000~\cite{ding2022event} & EventHDR\\ \hline
			Year	&2016&	2020	&2022&	2022&	2024 \\
			Task	&HDR Recon.&	HDR Recon.&	Video Interp.	&Video Deblur.	&HDR Recon. \\
			Train/Test	& Test	&Test &	Train \& Test&	Train \& Test	&Train \& Test \\
			HDR/LDR	&LDR&	LDR	&HDR&	LDR	&HDR \\
			Bit-Depth &	8	&8	&8&	8&	16 \\
			Frame Rate &	24 fps	&$\textless$40	fps&28	fps&2000	fps&500-2000 fps\\
			Num. Frames & 28418 & 15390 & ~40000 & 2565 & 81128\\
 			\hline\hline
		\end{tabular}
	\end{threeparttable}
\end{table*}

\section{Experiments}
In this section, we first provide details of our experimental settings. Then, qualitative and quantitative compared results are evaluated. After that, we extend our work to downstream applications on event-based vision. Finally, we conduct experiments to analyze the network architecture and data requirement for high-speed event-to-HDR tasks. 

\begin{figure*}[h!] \small
	\centering
	\setlength{\tabcolsep}{1pt}
	\footnotesize
	\begin{tabular}{cccccccccccc}
		& \multicolumn{3}{c}{Scene 1} & & \multicolumn{3}{c}{Scene 2} & & \multicolumn{3}{c}{Scene 3}\\
		\rotatebox{90}{\quad Low bits} &
		\includegraphics[width=.1\linewidth,clip,keepaspectratio]{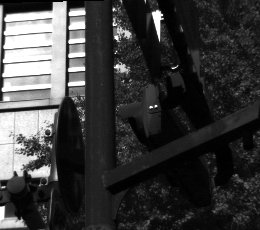} & \includegraphics[width=.1\linewidth,clip,keepaspectratio]{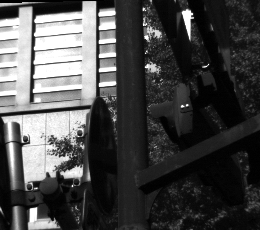} & \includegraphics[width=.1\linewidth,clip,keepaspectratio]{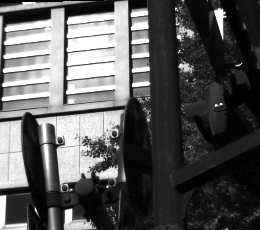} && \includegraphics[width=.1\linewidth,clip,keepaspectratio]{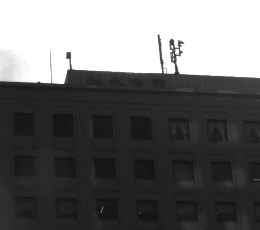} & \includegraphics[width=.1\linewidth,clip,keepaspectratio]{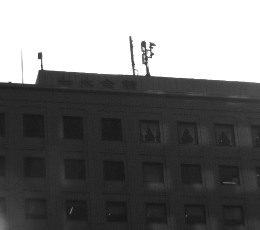} & \includegraphics[width=.1\linewidth,clip,keepaspectratio]{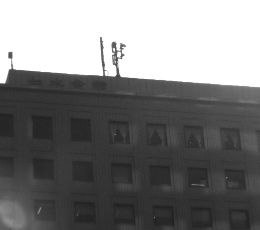} && \includegraphics[width=.1\linewidth,clip,keepaspectratio]{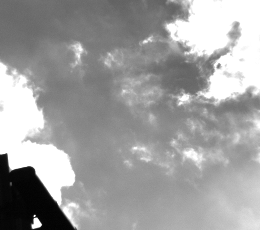} & \includegraphics[width=.1\linewidth,clip,keepaspectratio]{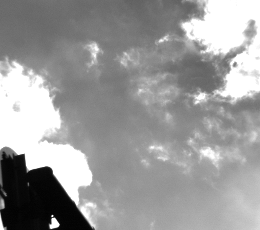} & \includegraphics[width=.1\linewidth,clip,keepaspectratio]{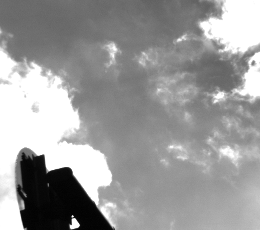}\\
		\rotatebox{90}{\quad High bits} &
		\includegraphics[width=.1\linewidth,clip,keepaspectratio]{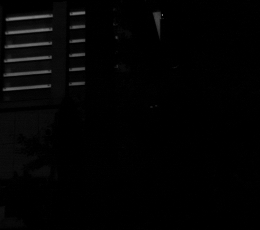} & \includegraphics[width=.1\linewidth,clip,keepaspectratio]{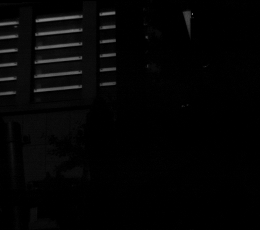} & \includegraphics[width=.1\linewidth,clip,keepaspectratio]{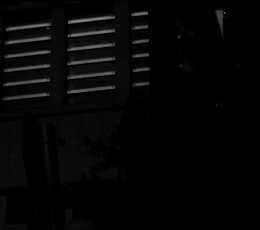} && \includegraphics[width=.1\linewidth,clip,keepaspectratio]{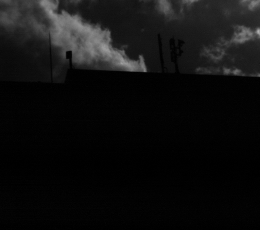} & \includegraphics[width=.1\linewidth,clip,keepaspectratio]{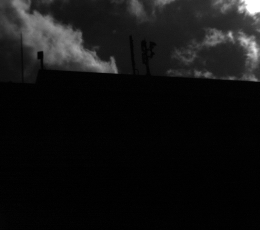} & \includegraphics[width=.1\linewidth,clip,keepaspectratio]{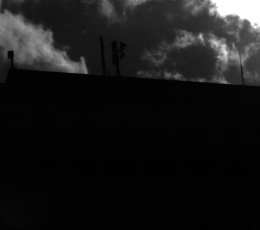} && \includegraphics[width=.1\linewidth,clip,keepaspectratio]{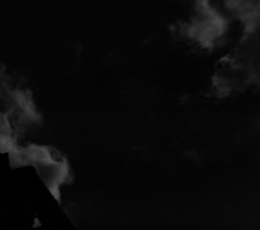} & \includegraphics[width=.1\linewidth,clip,keepaspectratio]{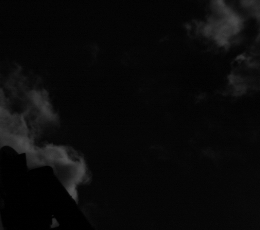} & \includegraphics[width=.1\linewidth,clip,keepaspectratio]{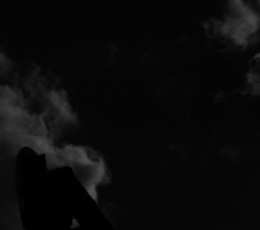}\\
		\rotatebox{90}{ \qquad HDR} &
		\includegraphics[width=.1\linewidth,clip,keepaspectratio]{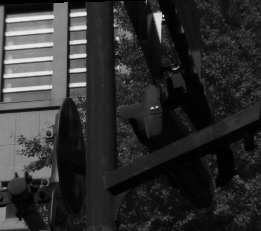} & \includegraphics[width=.1\linewidth,clip,keepaspectratio]{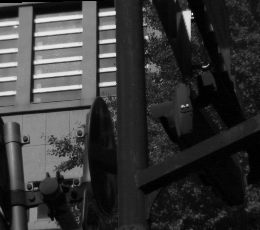} & \includegraphics[width=.1\linewidth,clip,keepaspectratio]{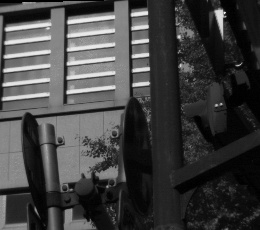} && \includegraphics[width=.1\linewidth,clip,keepaspectratio]{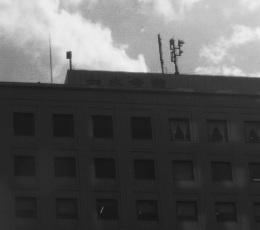} & \includegraphics[width=.1\linewidth,clip,keepaspectratio]{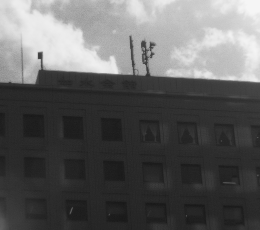} & \includegraphics[width=.1\linewidth,clip,keepaspectratio]{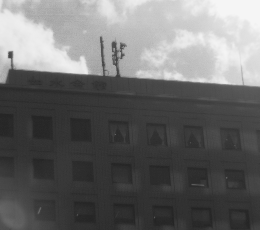} && \includegraphics[width=.1\linewidth,clip,keepaspectratio]{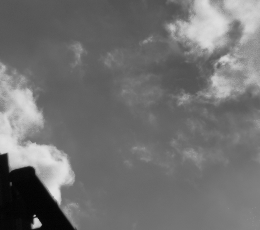} & \includegraphics[width=.1\linewidth,clip,keepaspectratio]{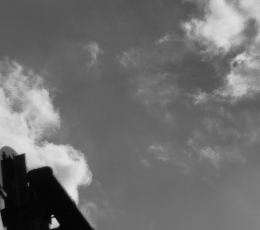} & \includegraphics[width=.1\linewidth,clip,keepaspectratio]{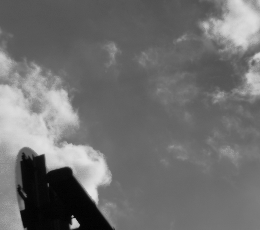}\\
		
		& Frame 1 & Frame 2 & Frame 3 & & Frame 1 & Frame 2 & Frame 3 & & Frame 1 & Frame 2 & Frame 3\\
	\end{tabular}
	\vspace{-2mm}
	\caption{Three representative scenes of our captured real dataset. In order to recognize the scene motions, the shown two consecutive frames are chosen at the interval of $100$ real frames.}
	\label{fig:real_data}
	\vspace{-2mm}
\end{figure*}

\subsection{Experimental Settings}
We compare our model with seven state-of-the-art Event-to-Video reconstruction methods, including: 1) A traditional non-deep method based on high-pass filter~\cite{scheerlinck2018continuous} (HF); 2) Two pioneering methods including a recurrent U-Net~\cite{rebecq2019high} (E2VID) and a fast reconstruction network \cite{scheerlinck2020fast} (FireNet); 3) Four more recent deep learning based methods including a model pretrained on a high quality event-to-frame dataset~\cite{stoffregen2020reducing} (E2VID+), an event super-resolution methods based on optical flow warping~\cite{mostafaviisfahani2021e2sri} (E2SRI), a transformer based event-to-video reconstruction model~\cite{weng2021event} (EITR), and a spiking neural network approach~\cite{zhu2022event} (EVSNN).
We reproduced these methods using publicly available codes and tested them on both simulated datasets and our real EventHDR dataset.

The reconstruction results for all methods are assessed using four image quality metrics: Root Mean Squared Error (RMSE), Structural Similarity~\cite{wang2004image} (SSIM), Learned Perceptual Image Patch Similarity~\cite{zhang2018unreasonable} (LPIPS), and temporal consistency loss (TC) introduced in Section ~\ref{sec:network}. While RMSE evaluates the overall prediction error, SSIM measures the 2D spatial fidelity. LPIPS gauges perceptual similarity, and TC quantifies the temporal continuity of an image sequence.

\subsection{Comparisons of State-of-the-Arts}
Here, we perform evaluations using all compared methods on a simulated event dataset. Then, we present experimental results on our EventHDR dataset to further validate the capability of our model in handling real-world scenes with challenging lighting conditions. This two-step evaluation process enables us to demonstrate the effectiveness of our method in both simulated and real scenarios, highlighting its potential in real applications.

\begin{figure}[tbp]
	\centering
	\setlength\tabcolsep{1pt}
	\footnotesize
	\resizebox{\linewidth}{!}{
		\begin{tabular}{ccccc}
			\includegraphics[width=0.19\linewidth]{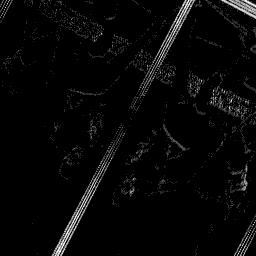}
			&\includegraphics[width=0.19\linewidth]{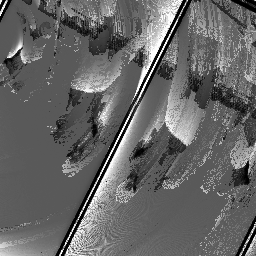}
			&\includegraphics[width=0.19\linewidth]{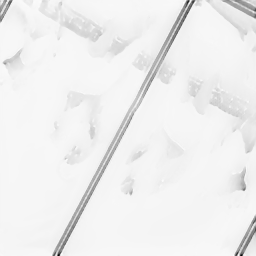}
			&\includegraphics[width=0.19\linewidth]{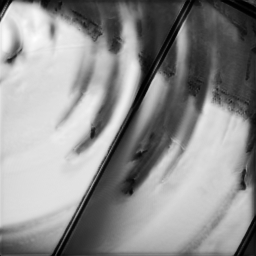}
			&\includegraphics[width=0.19\linewidth]{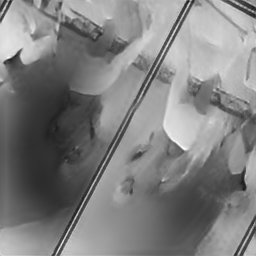} \\
			Event
			& HF~\cite{scheerlinck2018continuous}
			& E2VID~\cite{rebecq2019events}
			& FireNet~\cite{scheerlinck2020fast}
			& E2VID+~\cite{stoffregen2020reducing} \\
			\includegraphics[width=0.19\linewidth]{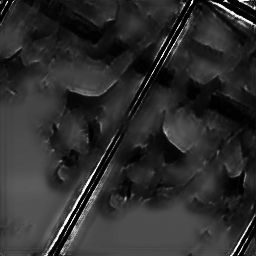}
			&\includegraphics[width=0.19\linewidth]{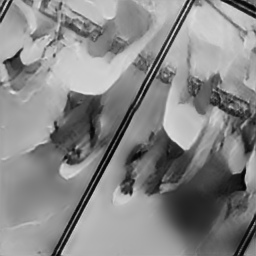}
			&\includegraphics[width=0.19\linewidth]{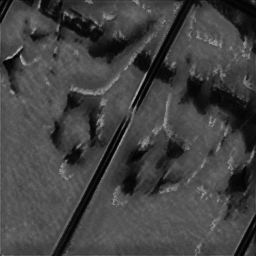}
			&\includegraphics[width=0.19\linewidth]{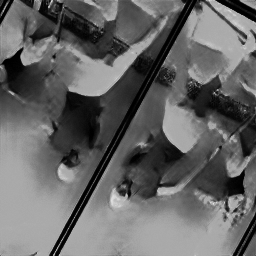}
			&\includegraphics[width=0.19\linewidth]{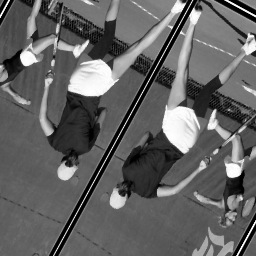}\\
			E2SRI~\cite{mostafaviisfahani2021e2sri}
			& EITR~\cite{weng2021event}
			& EVSNN~\cite{zhu2022event}
			& Ours
			& GT 
		\end{tabular}
	}
	\vspace{-2mm}
	\caption{Qualitative reconstruction results on simulated data.}
	\vspace{-3mm}
	\label{fig:performance_sim_sim}
\end{figure}

\begin{figure*}[tbp]
	\centering
	\footnotesize
	\setlength\tabcolsep{1pt}
	\resizebox{0.95\linewidth}{!}{
		\begin{tabular}{ccccc}
			\includegraphics[width=0.18\linewidth]{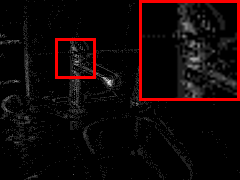}
			&\includegraphics[width=0.18\linewidth]{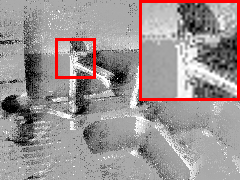}
			&\includegraphics[width=0.18\linewidth]{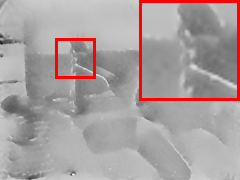}
			&\includegraphics[width=0.18\linewidth]{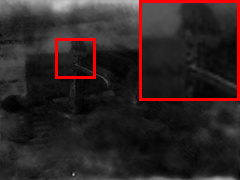}
			&\includegraphics[width=0.18\linewidth]{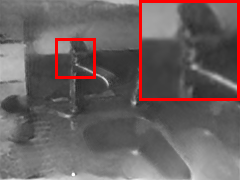} \\
			\vspace{-3.5mm}\\
			Event
			& HF~\cite{scheerlinck2018continuous}
			& E2VID~\cite{rebecq2019events}
			& FireNet~\cite{scheerlinck2020fast}
			& E2VID+~\cite{stoffregen2020reducing} \\
			\includegraphics[width=0.18\linewidth]{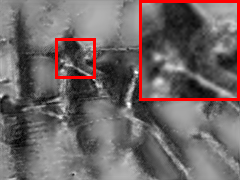}
			&\includegraphics[width=0.18\linewidth]{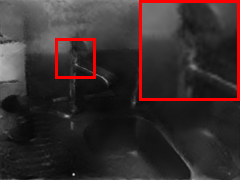}
			&\includegraphics[width=0.18\linewidth]{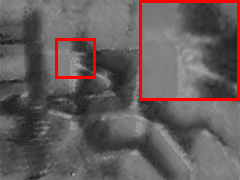}
			&\includegraphics[width=0.18\linewidth]{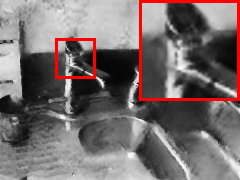}
			&\includegraphics[width=0.18\linewidth]{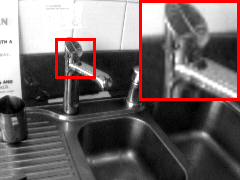}\\
			\vspace{-3.5mm}\\
			 E2SRI~\cite{mostafaviisfahani2021e2sri}
			 & EITR~\cite{weng2021event}
			 & EVSNN~\cite{zhu2022event}
			 & Ours
			 & GT \\
			\bottomrule
			\vspace{-2.5mm}\\
			\includegraphics[width=0.18\linewidth]{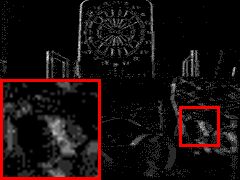}
			&\includegraphics[width=0.18\linewidth]{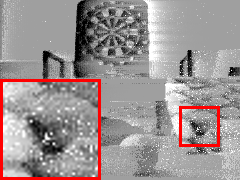}
			&\includegraphics[width=0.18\linewidth]{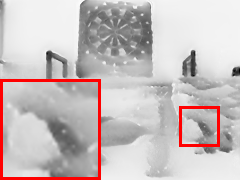}
			&\includegraphics[width=0.18\linewidth]{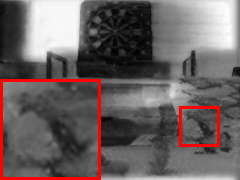}
			&\includegraphics[width=0.18\linewidth]{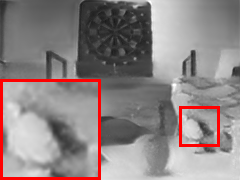} \\
			\vspace{-3.5mm}\\
			Event
			 & HF~\cite{scheerlinck2018continuous}
			 & E2VID~\cite{rebecq2019events}
			 & FireNet~\cite{scheerlinck2020fast}
			 & E2VID+~\cite{stoffregen2020reducing} \\
			\includegraphics[width=0.18\linewidth]{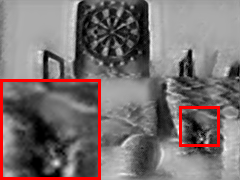}
			&\includegraphics[width=0.18\linewidth]{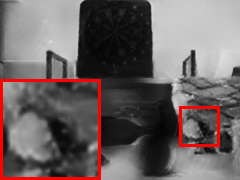}
			&\includegraphics[width=0.18\linewidth]{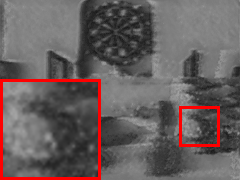}
			&\includegraphics[width=0.18\linewidth]{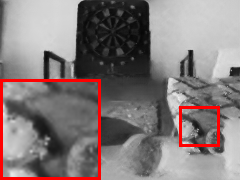}
			&\includegraphics[width=0.18\linewidth]{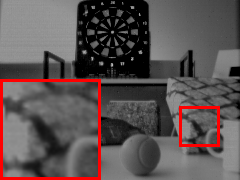}\\
			\vspace{-3.5mm}\\
			 E2SRI~\cite{mostafaviisfahani2021e2sri}
			 & EITR~\cite{weng2021event}
			 & EVSNN~\cite{zhu2022event}
			 & Ours
			 & GT \\
		\end{tabular}
	}
\caption{Qualitative reconstruction results on two public real evaluation dataset HQF~\cite{stoffregen2020reducing} and IJRR~\cite{mueggler2017event}. For each dataset, we provide the results of a typical scene for all compared methods.}
\label{fig:performance_sim_real}
\end{figure*}

\begin{figure*}[tbp]
	\centering
	\footnotesize
	\setlength\tabcolsep{1pt} 
	\resizebox{0.95\linewidth}{!}{
		\begin{tabular}{ccccc}
			\includegraphics[width=0.18\linewidth]{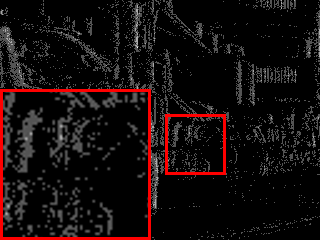}
			& \includegraphics[width=0.18\linewidth]{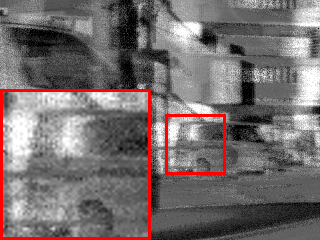}
			& \includegraphics[width=0.18\linewidth]{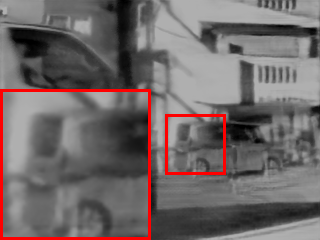}
			& \includegraphics[width=0.18\linewidth]{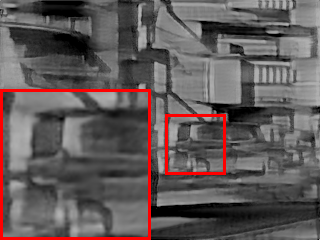}
			& \includegraphics[width=0.18\linewidth]{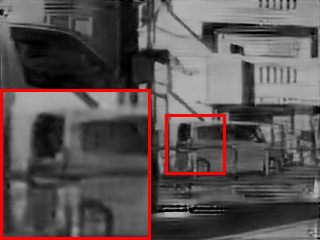} \\
			\vspace{-3.5mm} \\
			Event
			& HF~\cite{scheerlinck2018continuous}
			& E2VID~\cite{rebecq2019events}
			& FireNet~\cite{scheerlinck2020fast}
			& E2VID+~\cite{stoffregen2020reducing} \\
			\includegraphics[width=0.18\linewidth]{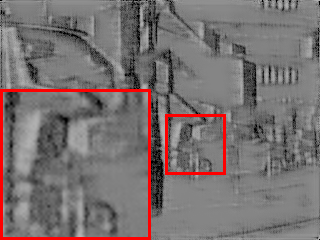}
			& \includegraphics[width=0.18\linewidth]{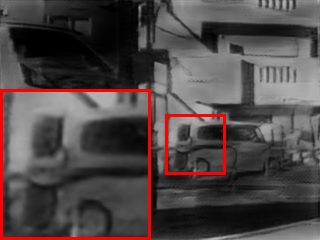}
			& \includegraphics[width=0.18\linewidth]{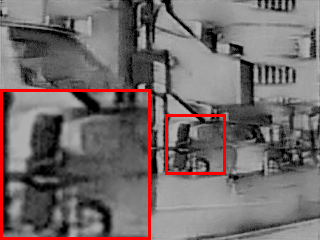}
			& \includegraphics[width=0.18\linewidth]{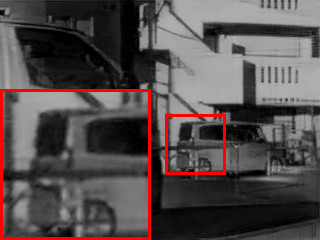}
			& \includegraphics[width=0.18\linewidth]{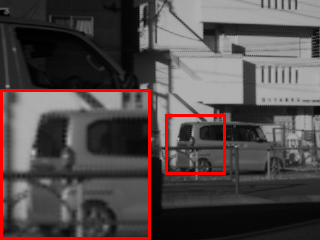} \\
			\vspace{-3.5mm} \\
			E2SRI~\cite{mostafaviisfahani2021e2sri}
			& EITR~\cite{weng2021event}
			& EVSNN~\cite{zhu2022event}
			& Ours
			& GT \\
			\bottomrule \\
			\vspace{-5.5mm} \\
			\includegraphics[width=0.18\linewidth]{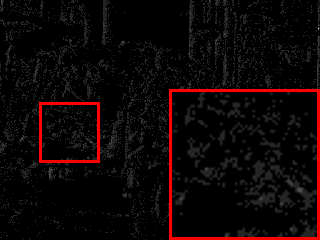}
			& \includegraphics[width=0.18\linewidth]{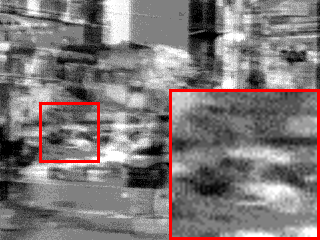}
			& \includegraphics[width=0.18\linewidth]{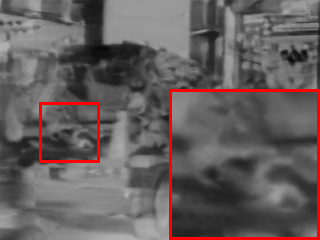}
			& \includegraphics[width=0.18\linewidth]{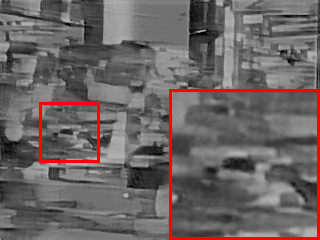}
			& \includegraphics[width=0.18\linewidth]{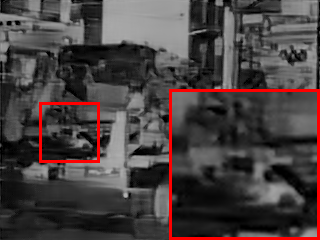} \\
			\vspace{-3.5mm} \\
			Event
			& HF~\cite{scheerlinck2018continuous}
			& E2VID~\cite{rebecq2019events}
			& FireNet~\cite{scheerlinck2020fast}
			& E2VID+~\cite{stoffregen2020reducing} \\
			\includegraphics[width=0.18\linewidth]{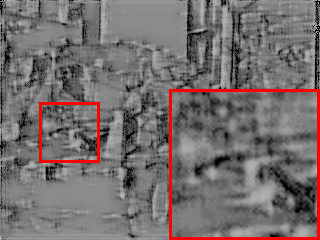}
			& \includegraphics[width=0.18\linewidth]{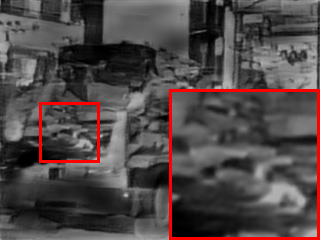}
			& \includegraphics[width=0.18\linewidth]{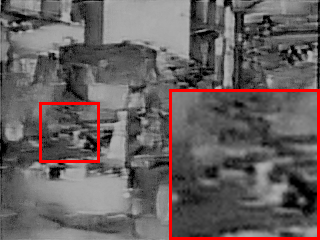}
			& \includegraphics[width=0.18\linewidth]{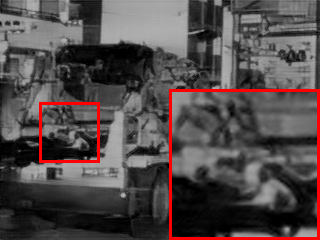}
			& \includegraphics[width=0.18\linewidth]{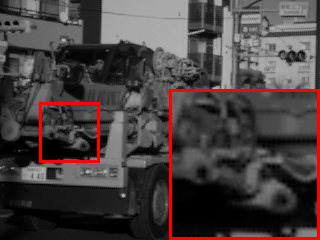} \\
			\vspace{-3.5mm} \\
			E2SRI~\cite{mostafaviisfahani2021e2sri}
			& EITR~\cite{weng2021event}
			& EVSNN~\cite{zhu2022event}
			& Ours
			& GT \\
	\end{tabular}}
	\caption{Qualitative reconstruction results on our real-world EventHDR data. We provide the results of two typical scenes for all compared methods.}
	\label{fig:performance_comparison_real}
	\vspace{-1mm}
\end{figure*}

\subsubsection{Performances on Simulated Data} 
\label{sec:ordinary_results}

Like previous methods~\cite{stoffregen2020reducing, rebecq2019events, rebecq2019high, weng2021event}, in the first case, we train all event-to-HDR video reconstruction networks on a synthetic dataset, which is simulated using ESIM simulator~\cite{rebecq2018esim}. We follow the pipeline of Stoffregen~\etal~\cite{stoffregen2020reducing}, which carefully analyze several significant factors to generate realistic simulated data by reducing the the gap between simulated and real data. In total, we generate $200$ paired event/video sequences using MS-COCO~\cite{lin2014microsoft} with $10$ seconds each. $160$ of them are randomly chosen for training, and the other sequences are used for evaluation.

Non-deep method HF~\cite{scheerlinck2018continuous} is directly evaluated on the $40$ testing sequences, while all deep learning-based methods~\cite{rebecq2019high, stoffregen2020reducing, mostafaviisfahani2021e2sri, weng2021event, zhu2022event, teng2022nest,scheerlinck2020fast} are trained with the same number of training iterations for fair comparisons.

Table~\ref{tab:performance_comparison_simulated} (Simulated Data) summarizes the numerical results based on the average of all metrics, with the best performance highlighted in \textbf{bold}. Among the compared methods, the traditional method HF cannot compete with other deep learning-based methods. Within the comparison among deep learning-based methods, E2VID+ shows better results than E2VID due to the pretraining on their high-quality synthetic data, even though they both use the same recurrent U-Net. In other words, the larger and more realistic data used for E2VID+ pretraining can indeed improve the reconstruction task.  EVSNN, which uses spiking neural networks, is lightweight but comes at the cost of reduced reconstruction precision. Our method provides better results for all error metrics, and the average results of all scenes significantly outperform the compared methods in both spatial and temporal domains, demonstrating the superiority of our proposed convolutional recurrent neural network.

To better illustrate the experimental results, several representative restored videos are shown in Fig. \ref{fig:performance_sim_sim}. From left to right, the displayed sequence includes the event frame followed by reconstructed frames from HF~\cite{scheerlinck2018continuous}, E2VID~\cite{rebecq2019events}, FireNet~\cite{scheerlinck2020fast}, E2VID+~\cite{stoffregen2020reducing}, E2SRI~\cite{mostafaviisfahani2021e2sri}, EITR~\cite{weng2021event}, EVSNN~\cite{zhu2022event}, our proposed method, and the ground truth. Notably, our method yields results that are more aligned with the ground truth, which is consistent with the numerical findings. This substantiates the superior efficacy of our approach in reconstructing HDR videos from event camera data in comparison to state-of-the-art methods.

\begin{table*}[h!]
	\centering
	\footnotesize
	\setlength{\tabcolsep}{2.5mm}
	\resizebox{0.98\linewidth}{!}{
	\begin{threeparttable}
		\caption{Evaluation results on simulated data and two public real dataset HQF~\cite{stoffregen2020reducing} and IJRR~\cite{mueggler2017event}. All methods are trained on the same synthetic dataset introduced in Section~\ref{sec:ordinary_results}.}
		\vspace{-2mm}
		\label{tab:performance_comparison_simulated}
		\begin{tabular}{cccccccccc}
			\hline\hline
			Scene&Metrics&HF~\cite{scheerlinck2018continuous}& E2VID~\cite{rebecq2019events}&FireNet~\cite{scheerlinck2020fast} &E2VID+~\cite{stoffregen2020reducing}&E2SRI~\cite{mostafaviisfahani2021e2sri}&EITR~\cite{weng2021event}&EVSNN~\cite{zhu2022event} & Ours\\
			\hline
			\multirow{4}{*}{Simulated Data}  
			& RMSE $\downarrow$ & 0.2874 & 0.3004  & 0.4303  &  0.3136  & 0.2709  & 0.2643  & 0.3292    & \textbf{0.2480}  \\ 
			& SSIM $\uparrow$ & 0.3136  & 0.3035  & 0.2539  & 0.3430  & 0.3728  & 0.4024  & 0.3197   & \textbf{0.4595}  \\ 
			& LPIPS $\downarrow$  & 0.6333  & 0.6442  & 0.6613  & 0.6018  & 0.5500  & 0.5409  & 0.5609   & \textbf{0.4983}  \\ 
			& TC $\downarrow$  & 0.2327  & 0.1434  & 0.1694  &  \textbf{0.1246}  & 0.4289  & 0.1926  & 0.2497   & 0.1316  \\ 
			\hline
			\multirow{4}{*}{HQF}  
			&RMSE $\downarrow$  & 0.3683 &0.3319  & 0.2800  & 0.2630  & 0.2746  & 0.2220  & 0.2765    & \textbf{0.2193}  \\ 
			&SSIM $\uparrow$ & 0.1877  &0.2680  & 0.3389  & 0.3927  & 0.3881  & 0.4240  & 0.3288   & \textbf{0.4825}  \\ 
			&LPIPS $\downarrow$  & 0.6595  &0.5827  & 0.5749  & 0.5346  & 0.5492  & 0.5057  & 0.5597  & \textbf{0.4645}  \\ 
			&TC $\downarrow$  & 0.6365 &0.3845  & 0.3983  & 0.3778  & 0.7828  & 0.4585  & 0.4696  & \textbf{0.3630}  \\ 
			\hline
			\multirow{4}{*}{IJRR}  
			&RMSE $\downarrow$  & 0.3449   & 0.4085  &0.2859  & 0.2397  & 0.2565  & 0.2624  & 0.2711   &\textbf{ 0.2028}  \\ 
			&SSIM  $\uparrow$& 0.1953   & 0.3484  &0.3722   & 0.4574  & 0.3980  & 0.4244  & 0.3224   & \textbf{0.5095}  \\ 
			&LPIPS $\downarrow$  & 0.6303   & 0.5864  & 0.4946 & 0.4715  & 0.5137  & 0.4575  & 0.5179   & \textbf{0.4233}  \\ 
			&TC $\downarrow$  & 0.5739   & 0.6341  & 0.4761  &0.5092  & 0.8347  & 0.5750  & 0.5888  & \textbf{0.4458}  \\ 

			\hline\hline
		\end{tabular}
		
	\end{threeparttable}}
\end{table*}

\subsubsection{Performances on Real Event Stream}
\label{sec:real_results}
Utilizing the model trained on simulated data in Section \ref{sec:ordinary_results}, we further follow \cite{stoffregen2020reducing} and \cite{weng2021event} to evaluate directly on event streams captured by real event cameras. The real event testing datasets include HQF~\cite{stoffregen2020reducing} and IJRR~\cite{mueggler2017event}.

The quantitative and qualitative results are displayed in Table~\ref{tab:performance_comparison_simulated} and Fig.~\ref{fig:performance_sim_real}. Our network still outperforms all compared methods. It is worth mentioning that although the visual results of simulated experiments shown in Section~\ref{sec:ordinary_results} appear pleasing, when it comes to real event streams evaluation, the reconstruction results of all methods share a common issue, \ie, despite preserving the details of HDR scenes, the reconstruction results seem unnatural and differ significantly from human visual perception. This issue arises due to the domain gap between simulated training data and real evaluation data. In other words, \textit{the practicality of previous event-to-HDR methods E2VID~\cite{rebecq2019high} , E2VID+~\cite{stoffregen2020reducing} and EITR~\cite{weng2021event} are limited by their synthetic training data}.

\subsubsection{Experiments on Our EventHDR Dataset}
\label{sec:hdr_results}

To explore the potential of event cameras for real-world HDR imaging, we conduct comparative studies of our method against existing approaches, utilizing our captured real-world dataset. For a fair evaluation, we use the publicly available code of these method and retrained them on our EventHDR training set. Additionally, to ensure consistency, the number of training iterations was kept the same across all methods. The average quantitative results are presented in Table \ref{tab:performance_comparison_real}. Our method surpasses all competing methods in both spatial and temporal metrics, aligning with the outcomes of experiments conducted on simulated data in Sections~\ref{sec:ordinary_results} and \ref{sec:real_results}.

To visualize the results, several representative reconstructed frames are shown in Fig. \ref{fig:performance_comparison_real}. The event frame and representative reconstructed frames of HF~\cite{scheerlinck2018continuous}, E2VID~\cite{rebecq2019events}, FireNet~\cite{scheerlinck2020fast}, E2VID+~\cite{stoffregen2020reducing}, E2SRI~\cite{mostafaviisfahani2021e2sri}, EITR~\cite{weng2021event}, EVSNN~\cite{zhu2022event}, Ours, and ground truth are shown from left to right. We observe that the frames recovered by our method closely resemble the ground truth and significantly outperform other reconstruction methods. Notably, as mentioned in Section~\ref{sec:real_results}, although the previous strategy of training the reconstruction models on simulated data can be directly applied to real event streams, the results shown in  Fig.~\ref{fig:performance_comparison_real} are much more visually pleasing than the reconstruction results shown in Fig.~\ref{fig:performance_sim_real}. \textit{This improvement is attributed to our EventHDR dataset that contains real paired event and HDR training sets, which helps avoid the domain gap that previous methods E2VID~\cite{rebecq2019high}, E2VID+~\cite{stoffregen2020reducing} and EITR~\cite{weng2021event} suffer from}.

\begin{table*}[h!]
	\renewcommand{\arraystretch}{1.}
	\centering
	\vspace{-2mm}
	\footnotesize
	\setlength{\tabcolsep}{3.5mm}
	\resizebox{0.98\linewidth}{!}{
	\begin{threeparttable}
		\caption{Performance comparisons on real data, along with the computational costs (Computed at $128 \times 128$ resolution). All methods are trained and evaluated on our EventHDR dataset.}
		\label{tab:performance_comparison_real}
		\begin{tabular}{ccccccccc}
			\hline\hline
			Metrics&HF~\cite{scheerlinck2018continuous}& E2VID~\cite{rebecq2019events}& FireNet~\cite{scheerlinck2020fast} & E2VID+~\cite{stoffregen2020reducing}&E2SRI~\cite{mostafaviisfahani2021e2sri}&EITR~\cite{weng2021event}&EVSNN~\cite{zhu2022event}& Ours\\
			\hline
			RMSE $\downarrow$ & 0.3792 & 0.1794  & 0.2337  & 0.1621  & 0.3964  & 0.1156  & 0.4201   & \textbf{0.1039}  \\ 
			SSIM  $\uparrow$& 0.1283 & 0.3576  & 0.2466  & 0.4954  & 0.1719  & 0.5762  & 0.1519   & \textbf{0.5975}  \\ 
			LPIPS$\downarrow$  & 0.7665 & 0.1829  &  0.2199  & 0.4872  & 0.2493  & 0.2586  & 0.2336  & \textbf{0.1570}  \\ 
			TC $\downarrow$ & 0.2944  & 0.3024  & 0.2646  & 0.4020  & 0.6880  & 0.3066  & 0.4162  &\textbf{ 0.2524}  \\
			\hline
			Params (M) & / & 10.71 & 0.04 & 10.71 & 7.63 & 22.18 & 4.41 & 4.44 \\
			Macs (G) & / & 7.51 & 0.63 & 7.51 & 249.79 & 11.15 & 59.83 & 76.91 \\
			\hline\hline
		\end{tabular}
		\vspace{-2mm}
	\end{threeparttable}}
\end{table*}

\subsection{Ablation Study on Network Design}
Here, we aim to thoroughly analyze the network design for event-to-HDR tasks, to verify the effectiveness of our reconstruction model. More importantly, we seek to identify suitable network components for the specific event-to-HDR task, which may guide other researchers in designing network architecture for this task.

\subsubsection{Propagation Module}
To process video-like sequences, the input data can be generally propagated through three different ways. First, the encoder works in a sliding-window fashion and takes neighboring frames as input to reconstruct the image at the central timestamp~\cite{wang2019edvr, tassano2020fastdvdnet, yue2020supervised}. For the second approach, sequences can be processed in a recurrent manner through a uni-directional RNN architecture~\cite{zhong2020efficient, rebecq2019events}, thus utilizing the temporal similarity during the recurrent propagation. A third method for feature propagation is the bi-directional RNN~\cite{chan2021basicvsr, chan2022basicvsrplus}, which is developed from the uni-directional RNN and further incorporates information from both the past and future states.

We compare the three approaches by simply modifying the proposed network in Section~\ref{sec:method}. In addition, we also add the proposed key frame guidance (KFG) strategy to assist the learning of the uni-directional RNN. The numeric results are shown in Table~\ref{tab:prop}, and some typical visual results are shown in Fig.~\ref{fig:ablation_recurrent}. Notably, compared with the three recurrent networks, we observe that the sliding-window propagation with only neighboring frames appears to have severe dark artifacts, which also greatly affect the quantitative results. This phenomenon can be intuitively understood, considering the sparse attribute of event streams, for event cameras only record difference information. The uni-directional RNN also suffers from information loss. After a careful inspection of the results, we find that long-range error propagation is the main cause. By incorporating the KFG module, the error propagated through hidden states is refreshed and leads to the best performance. Although the bi-directional RNN also presents competitive results, the need for future frames limits its practicality in real applications, \textit{i.e.}, real-time HDR reconstruction.

\begin{table}
	\centering
	\vspace{-2mm}
	\footnotesize
	\caption{Ablation study on feature propagation module. We provide the results for  sliding-window (SW), uni-directional (Uni.), bi-directional (Bi.), and uni-directional with key frame guidance (Uni. + KFG).}\label{tab:prop}
	\vspace{-2mm}
	\setlength{\tabcolsep}{3mm}
	\renewcommand{\arraystretch}{1}
	\begin{tabular}{ccccc}
		\hline\hline
		Settings & RMSE $\downarrow$ & SSIM $\uparrow$ & LPIPS$\downarrow$  & TC$\downarrow$    \\
		\hline
		SW  & 0.1153 	& 	0.5574 &	0.1642 &	0.3728 \\
		Uni-directional   &  0.1105 & 0.5563 & 0.1621 & 0.3125\\ 
		Bi-directional   & 0.1046 & \textbf{0.5993} & 0.1584 & 0.2563\\
		Uni. + KFG   & \textbf{0.1039} & 0.5975 & \textbf{0.1570} & \textbf{0.2524}  \\
		\hline\hline
	\end{tabular}
		\vspace{-1mm}
\end{table}

To further refine the application of the KFG module within our uni-directional RNN architecture, we conduct an ablation study focusing on the selection of key frame intervals. This study involves adjusting the interval $K$, while maintaining other settings constant. We set $K$ to 1, 2, 5, 10, and obtain the quantitative results presented in Table~\ref{tab:keyframe}. We find that though frequent key frames ($K=1, 2$) provide promising reconstruction at the cost of larger computational burden. When increasing the interval to more than 5 frames, a performance degradation  appears, possibly due to the loss of information during longer passes. Consequently, we choose $K=5$ for both effectiveness and efficiency.

\begin{table}
	\centering
	\caption{Ablation study on the interval of key frames.}\label{tab:keyframe}
	\vspace{-2mm}
	\setlength{\tabcolsep}{5mm}
	\renewcommand{\arraystretch}{1}
	\footnotesize
	\begin{tabular}{ccccc}
		\hline\hline
		K & RMSE $\downarrow$ & SSIM $\uparrow$ & LPIPS$\downarrow$  & TC$\downarrow$     \\
		\hline
		1   & 0.1118  &	0.5972 &	0.1623 & 0.3072   \\
		2 	&0.1045	&0.5956	&\textbf{0.1568}	&0.2543 \\
		5   &\textbf{0.1039} & \textbf{0.5975} & 0.1570 & \textbf{0.2524}  \\ 
		10 &0.1132	&0.5647	&0.1634&	0.2531\\
		\hline\hline
	\end{tabular}
\end{table}

\begin{figure}[t] \small
	\centering
	\setlength{\tabcolsep}{1pt}
	\begin{tabular}{ccc}
		
		\includegraphics[width=.33\linewidth,clip,keepaspectratio]{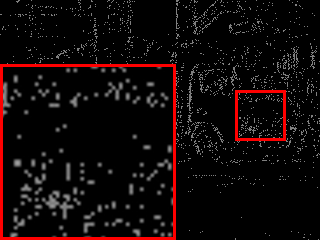} & 
		\includegraphics[width=.33\linewidth,clip,keepaspectratio]{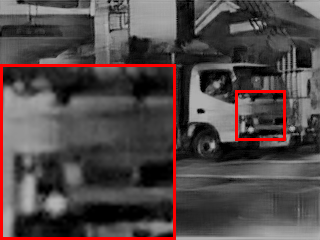} & 
		\includegraphics[width=.33\linewidth,clip,keepaspectratio]{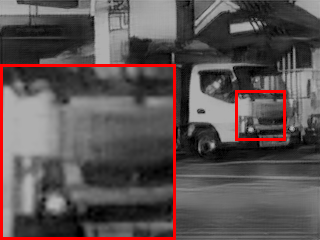} \\
		Event Frame & SW & Uni-directional \\\vspace{-2mm}\\
		\includegraphics[width=.33\linewidth,clip,keepaspectratio]{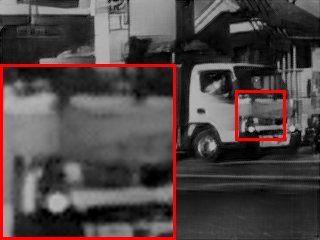} & 
		\includegraphics[width=.33\linewidth,clip,keepaspectratio]{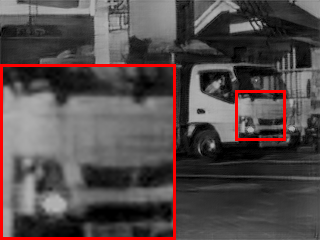} & 
		\includegraphics[width=.33\linewidth,clip,keepaspectratio]{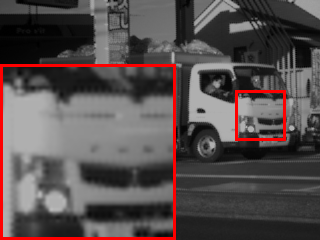} \\
		Bi-directional & Uni. + KFG & GT
		
	\end{tabular}
	\caption{Visual results for our network with different propagation methods, including sliding-window (SW), uni-directional (Uni.), bi-directional (Bi.), and uni-directional with key frame guidance (Uni. + KFG).}
	\vspace{-2mm}
	\label{fig:ablation_recurrent}
\end{figure}

\subsubsection{Alignment Module}
Proper alignment of neighboring input frames is crucial for effective fusion in event-to-HDR tasks. Here, we analyze the impact of three alignment methods: no alignment, optical flow alignment, and the PCD alignment we use in our network. For optical flow estimation, we use SpyNet~\cite{ranjan2017optical}, which is computationally efficient and can provide a reasonable alignment performance. 

We compare the three alignment methods by modifying the proposed network in Section~\ref{sec:method} accordingly. The numeric results are shown in Table~\ref{tab:align}. We observe that neglecting alignment significantly degrades performance, highlighting the importance of alignment for event-to-HDR reconstruction. While optical flow  offers a noticeable improvement over the non-alignment approach, it does not achieve optimal results due to possible estimation inaccuracies. Contrastingly, the PCD alignment strategy we employ emerges as the most effective, as it effectively captures complex motions and large displacements, resulting in a more accurate reconstruction. This demonstrates the effectiveness of the PCD alignment method in addressing the challenges posed by the event-to-HDR task.

\begin{table}[t]
	\centering
	\footnotesize
	\caption{Ablation study on different alignment components. We provide the reconstruction performances, computation complexity and  parameter counts of the alignment module.}\label{tab:align}
	\vspace{-2mm}
	\setlength{\tabcolsep}{0.8mm}
	\renewcommand{\arraystretch}{1}
	
	\begin{tabular}{ccccccc}
		\hline\hline
		Settings & RMSE $\downarrow$ & SSIM $\uparrow$ & LPIPS$\downarrow$  & TC$\downarrow$ & Params& Macs  \\
		\hline
		 No Alignment  &  0.1382 & 0.4657 &	0.1742 & 0.2569 & 0 & 0 \\
		Optical Flow   &  0.1279 & 0.4510 &	0.1802 &	0.2877 & 1.44M & 15.74G  \\ 
		PCD Alignment  &\textbf{0.1039} & \textbf{0.5975} & \textbf{0.1570} & \textbf{0.2524} & 1.24M & 35.85G  \\
		\hline\hline
	\end{tabular}
	\vspace{-1mm}
\end{table}

\subsubsection{Fusion Module}
Following the alignment of neighboring frame features, the next critical step is feature fusion. In our approach, we employ a local attention module to integrate multiple frame features effectively. Here, we provide  ablation studies to explore various feature fusion approaches.

We evaluate four distinct fusion approaches: simple addition, two convolutional layers, global temporal-spatial attention fusion mechanism, and our local attention fusion. To ensure a fair comparison, the network remains unchanged except for the fusion modules.

The quantitative results and computational efficiency on EventHDR dataset are summarized in Table~\ref{tab:fusion}. The results reveal a notable performance degradation with the simple addition and convolutional fusions, where no attention mechanisms are involved. This confirms the key role of attention blocks in effectively capturing and utilizing temporal and spatial information for event reconstruction tasks. In contrast, when comparing our local attention method with the global temporal-spatial attention scheme~\cite{wang2019edvr}, our approach not only achieves superior quantitative performance but also offers a reduction in both parameter count and computational operations. This efficiency is particularly valuable given the high-speed, sparse feature of event data, which demands efficient processing capabilities.

\begin{table}
	\centering
	\footnotesize
	\caption{Ablation study on fusion modules. We provide the reconstruction performances, computation complexity and  parameter counts of the several fusion approarches.}\label{tab:fusion}
	\vspace{-2mm}
	\setlength{\tabcolsep}{0.8mm}
	\renewcommand{\arraystretch}{1}
	\begin{tabular}{ccccccc}
		\hline\hline
		Settings & RMSE $\downarrow$ & SSIM $\uparrow$ & LPIPS$\downarrow$  & TC$\downarrow$ & Params& Macs  \\
		\hline
		Addition & 0.2159 &  0.3047 &	0.2221	& 0.3518 &0 & 0 \\
		Conv. Layers & 0.2206	& 0.3019 &	0.1935 &	0.2723 &12.35K &	0.20G	 \\
		Temporal-Spatial & 0.1063	& 0.5944 &	\textbf{0.1541} &	\textbf{0.2517} &	308.03K	& 3.91G \\
		Local Attention  &\textbf{0.1039} & \textbf{0.5975} & 0.1570 & 0.2524 & 40.18K	& 0.66G  \\
		\hline\hline
	\end{tabular}
\end{table}

\subsubsection{Loss Functions}
In this ablation study, we assess the influence of temporal consistency loss on the performance of event-to-HDR reconstruction. We compare two configurations, including our method without the temporal consistency loss and our complete model. The corresponding results are provided in Table~\ref{tab:temporal}. From the results, we observe that our complete model outperforms the configuration without temporal consistency loss in most metrics. This finding confirms the significance of the temporal consistency loss in enhancing temporal fidelity. Furthermore, it highlights the effectiveness of our deep recurrent reconstruction model in achieving superior performance for event-to-HDR tasks by incorporating the temporal consistency loss.

We further examined the effect of LPIPS loss on event-to-HDR reconstruction. The results in Table~\ref{tab:temporal} indicate that models lacking LPIPS failed to converge effectively. This is attributed to pixel-wise losses like $l_1$ being overly restrictive for the sparse and imprecise nature of event data. Incorporating LPIPS is crucial for successful training and convergence of our network, providing a loss function that better suits the characteristics of event-to-HDR tasks.
We further examined the effect of reconstruction loss on event-to-HDR reconstruction, specifically LPIPS and l1 losses. The results shown in Table~\ref{tab:temporal} demonstrate that models lacking LPIPS fail to converge effectively. This is due to pixel-wise losses like $l_1$ being overly restrictive given the sparse nature of event data. Incorporating LPIPS is crucial for successful training and convergence of our network, as it provides a loss function better suited to the characteristics of event-to-HDR tasks. While removing $l_1$ loss leads to some degradation in spatial fidelity and structural similarity, it suggests that although pixel-wise fidelity is not the primary driver of performance, it remains necessary.

\begin{table}[t]
	\centering
	\caption{Ablation study on loss functions.}\label{tab:temporal}
	\vspace{-2mm}
	\setlength{\tabcolsep}{3mm}
	\renewcommand{\arraystretch}{1}
	\begin{tabular}{ccccc}
		\hline\hline
		Settings & RMSE $\downarrow$ & SSIM $\uparrow$ & LPIPS$\downarrow$  & TC$\downarrow$     \\
		\hline
		w/o TC loss   & 0.1118  &	0.5972 &	0.1623 & 0.3072   \\
		w/o LPIPS loss   & 0.4536&	0.2045	&0.3563&	\textbf{0.2241}   \\
		w/o $l_1$ loss &0.1294&0.4260&0.2048&0.2703\\
		Ours   &\textbf{0.1039} & \textbf{0.5975} & \textbf{0.1570} &0.2524  \\ 
		\hline\hline
	\end{tabular}
\end{table}

\subsection{Explorations for Cross-Dataset Reconstruction}
In Section \ref{sec:real_results}, we demonstrate that the previous pipeline~\cite{stoffregen2020reducing, weng2021event, rebecq2019high, rebecq2019events}, which directly evaluates real HDR scenes using models trained on simulated data, has difficulty producing human-perception-like visualizations. To address this issue, we use the model pretrained on our EventHDR dataset to perform cross-camera HDR reconstruction on other real event evaluation sets, including HQF~\cite{stoffregen2020reducing} and IJRR~\cite{mueggler2017event}.

\begin{figure}[!t] \small
	\centering
	\setlength{\tabcolsep}{1pt}
	\begin{tabular}{cccc}
		\includegraphics[width=.24\linewidth,clip,keepaspectratio]{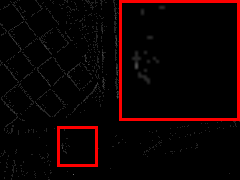} & 
		\includegraphics[width=.24\linewidth,clip,keepaspectratio]{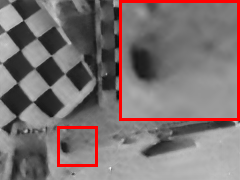} & 
		\includegraphics[width=.24\linewidth,clip,keepaspectratio]{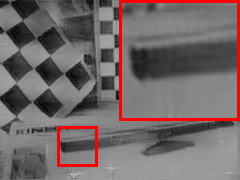} &
		\includegraphics[width=.24\linewidth,clip,keepaspectratio]{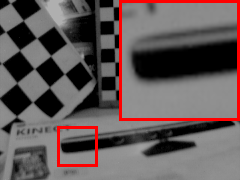} \\ 
		\includegraphics[width=.24\linewidth,clip,keepaspectratio]{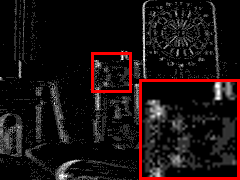} & 
		\includegraphics[width=.24\linewidth,clip,keepaspectratio]{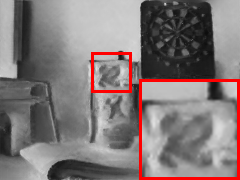} &
		\includegraphics[width=.24\linewidth,clip,keepaspectratio]{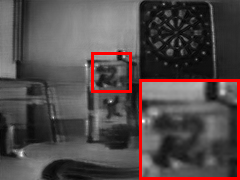} & 
		\includegraphics[width=.24\linewidth,clip,keepaspectratio]{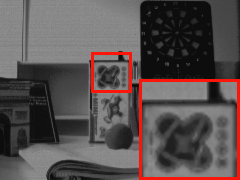} \\
		Event Frame & Simulated & EventHDR & GT
	\end{tabular}\vspace{-1mm}
	\caption{The experimental results for cross-camera and dataset evaluation. We provide the visual results produced by models trained on simulated data against those trained on our EventHDR data. The two scenes come from HQF~\cite{stoffregen2020reducing} and IJRR~\cite{mueggler2017event} datasets.}
	\vspace{-1mm}
	\label{fig:crosscamera}
\end{figure}

\begin{table}[t]
	\centering
	\caption{Evaluation results training on different data sources.}
	\vspace{-2mm}
	\setlength{\tabcolsep}{4mm}
	\renewcommand{\arraystretch}{1}
	\begin{threeparttable}
		\begin{tabular}{ccccc}
			\hline\hline
			Model & RMSE $\downarrow$ & SSIM $\uparrow$ & LPIPS$\downarrow$  & TC$\downarrow$    \\
			\hline
			Case 1 & 0.2032 &0.3191 	&0.2211 &	\textbf{0.2515} \\
			Case 2& 0.1231 	& 	0.5848 &	0.1601 &	0.2767 \\
			Case 3 & 0.1815 &	0.3163 &	0.3362 &	0.2656 \\
			Case 4 &  \textbf{0.1039} & \textbf{0.5975} & \textbf{0.1570} & 0.2524  \\ 
			
			\hline\hline
		\end{tabular}
		
		\label{tab:ablation_dataset}
	\end{threeparttable}
\end{table}

As shown in Fig.~\ref{fig:crosscamera}, when comparing the results of our network trained on the simulated dataset with those trained on our EventHDR dataset, we find that our dataset leads to more visually pleasing results while maintaining both the darkest and brightest regions of high dynamic scenes. This observation further confirms that our EventHDR dataset is effective in reconstructing comfortable HDR images and can be conveniently used for cross-camera and cross-dataset evaluations. As a result, our\textit{ EventHDR dataset has high applicability and practicality, facilitating the reconstruction of events captured by various individuals or camera brands}.

\subsection{Discussions on Event-to-HDR Training Data}
In this section, we delve into different types of training data for high-speed HDR video reconstruction from events, and evaluate on our real EventHDR testing data. We mainly consider four representative cases of training data, which are variants of our EventHDR dataset. These cases include
\begin{enumerate}
	\item Case 1: Input events are simulated from real LDR videos, and the LDR videos serve as ground truth. This case represents the training pipeline for previous event-to-HDR methods~\cite{stoffregen2020reducing, weng2021event, rebecq2019high, rebecq2019events}.
	\item Case 2: Input events are captured by real event camera, while LDR videos serve as ground truth. This case is similar to the paired event/APS training~\cite{wang2020eventsr}.
	\item Case 3: Real HDR videos serve as ground truth and are used to simulate input events.
	\item Case 4: Our EventHDR dataset, which consists of paired real high-speed event/HDR datasets.
\end{enumerate}

\begin{figure*}[!t] \small
	\centering
	\footnotesize
	\setlength{\tabcolsep}{1pt}
	\begin{tabular}{cccccc}
		\includegraphics[width=.16\linewidth,clip,keepaspectratio]{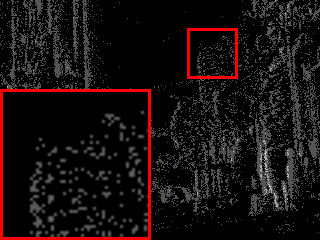}  & 
		\includegraphics[width=.16\linewidth,clip,keepaspectratio]{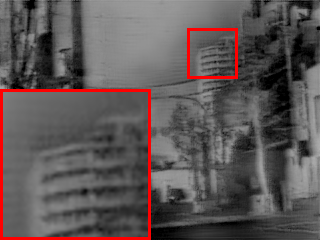} & 
		\includegraphics[width=.16\linewidth,clip,keepaspectratio]{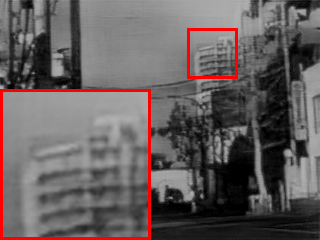} &
		\includegraphics[width=.16\linewidth,clip,keepaspectratio]{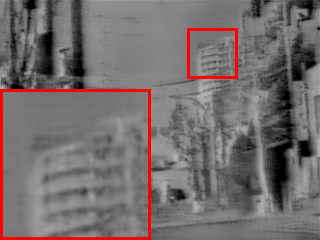} &
		\includegraphics[width=.16\linewidth,clip,keepaspectratio]{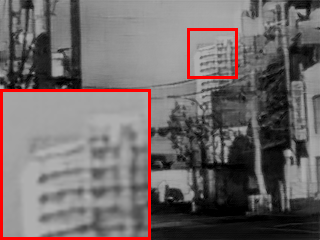} &
		\includegraphics[width=.16\linewidth,clip,keepaspectratio]{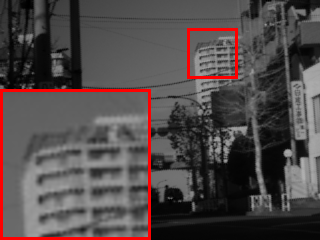}\\ 
		\includegraphics[width=.16\linewidth,clip,keepaspectratio]{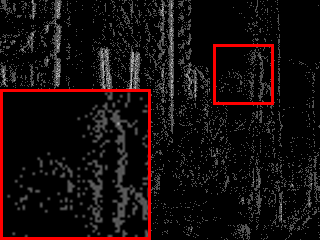}  & 
		\includegraphics[width=.16\linewidth,clip,keepaspectratio]{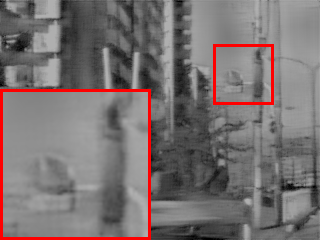} & 
		\includegraphics[width=.16\linewidth,clip,keepaspectratio]{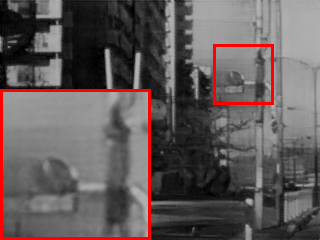} &
		\includegraphics[width=.16\linewidth,clip,keepaspectratio]{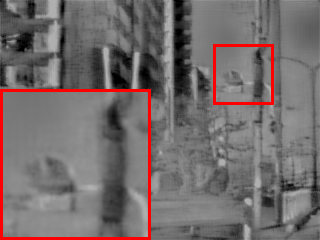} &
		\includegraphics[width=.16\linewidth,clip,keepaspectratio]{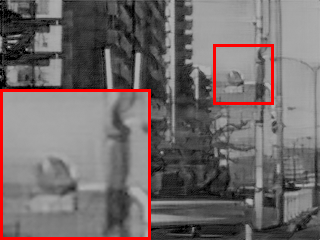} &
		\includegraphics[width=.16\linewidth,clip,keepaspectratio]{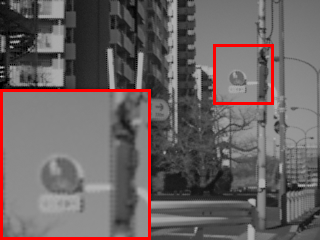}\\ 
		Event Frame & Case 1 & Case 2 & Case 3 & Case 4 & GT 
	\end{tabular}
	\vspace{-1mm}
	\caption{The HDR video reconstruction using different types of input data.}
	\vspace{-2mm}
	\label{fig:data_comparison}
\end{figure*}

We provide quantitative and qualitative results in Table~\ref{tab:ablation_dataset} and Fig.~\ref{fig:data_comparison}. By comparing all cases, it is evident that Case 1 performs the worst, primarily due to the domain gap between synthetic training data and real evaluation data. This observation verify the need for capturing high-quality real paired datasets for high-speed HDR video reconstruction from events. \textit{However, Case 1 is widely used in previous methods\cite{stoffregen2020reducing, weng2021event, rebecq2019high, rebecq2019events}, as capturing real paired training data like  EventHDR relies on intricate imaging system design and laborious data acquisition}. Furthermore, Cases 2 and 3 are improved variants from Case 1, which introduce HDR features during simulation, but the results remain unsatisfactory. As for our EventHDR dataset in Case 4, the unpleasant artifacts in Fig.~\ref{fig:data_comparison} are almost mitigated. In conclusion, our dataset addresses the key issues and enables better high-speed HDR reconstruction performance.

\subsection{Downstream HDR Applications}

Some methods intend to enhance event data in the stream domain~\cite{duan2021eventzoom,guo2022low}, and then apply applications directly on stream-like data~\cite{barranco2018real, gallego2018unifying}. However, a more natural and flexible pipeline~\cite{wang2021joint, mostafavi2021learning} for event-based HDR applications is to use an event camera to capture HDR event streams, then conduct event-to-HDR reconstruction methods, and finally apply frame-based vision algorithms~\cite{chen2023instance, fu2023category,fu2021coded} on the reconstructed HDR frames. In this way, the captured event streams can directly utilize existing well-established downstream vision algorithms. To further leverage the unique HDR capabilities of event cameras, we conduct extensive experiments on various downstream vision tasks. Specifically, we apply pretrained vision algorithms directly to the HDR images reconstructed from our EventHDR dataset, as discussed in Section~\ref{sec:hdr_results}. These experiments serve to demonstrate the advantages of our method and dataset. We perform four key downstream vision tasks, including object detection, panoptic segmentation, optical flow estimation, and monocular depth estimation.

To ensure a comprehensive and thorough comparison, we carry out successive high-level vision tasks on different types of images, including conventional intensity images, HDR reconstruction from event-to-HDR methods, and real HDR images. Specifically, the input images include

\begin{enumerate}
	\item Conventional intensity images captured by the active pixel sensor (APS), which is integrated with the event camera and captured along with the event streams, but is limited to a narrow dynamic range.
	\item Reconstructed video generated by our network, trained on the simulated data detailed in Section \ref{sec:ordinary_results} and following the pipeline established by \cite{stoffregen2020reducing}. This is denoted as Ours-sim.
	\item Reconstructed HDR video produced by several event-to-HDR methods trained on our real HDR dataset. We choose three most competitive methods according to Table \ref{tab:performance_comparison_real}, \ie, E2VID~\cite{rebecq2019events}, E2VID+~\cite{stoffregen2020reducing}, EITR~\cite{weng2021event}, and our method. We also incorporate a recent pretrained model HyperE2V~\cite{ercan2024hypere2vid}.
	\item The tone mapped ground truth HDR video, which contains enough details in both dark and bright regions, and can serves as a good baseline.
\end{enumerate}

\begin{table*}[t]
	\renewcommand{\arraystretch}{1.}
	\centering
	\footnotesize
	\caption{Quantitative results of downstream applications on our EventHDR dataset. We provide quantitative results for object detection, panoptic segmentation, optical flow estimation and monochrome depth estimation. The results of real HDR videos are used as ground truth for calculating evaluation metrics. The best performances are denoted in \textbf{bold}.}
	\vspace{-2mm}
	\begin{threeparttable}
		\begin{tabular}{p{2.7cm}<{\centering}p{1.35cm}<{\centering}p{1.35cm}<{\centering}p{1.35cm}<{\centering}p{1.45cm}<{\centering}p{1.45cm}<{\centering}p{1.45cm}<{\centering}p{1.45cm}<{\centering}p{1.45cm}<{\centering}}
			\hline\hline
			\multirow{2}{*}{Tasks} & \multirow{2}{*}{Metrics} & \multirow{2}{*}{APS} & \multirow{2}{*}{E2VID~\cite{rebecq2019events}} & \multirow{2}{*}{E2VID+~\cite{stoffregen2020reducing}} & \multirow{2}{*}{EITR~\cite{weng2021event}}  & \multirow{2}{*}{HyperE2V\cite{ercan2024hypere2vid}} & \multirow{2}{*}{Ours-sim} & \multirow{2}{*}{Ours} \\\\
			\hline
			
			\multirow{2}{*}{Object Detection} & Num.$\uparrow$ & 276	& 257	& 280& 	249	& 265& 188	&\textbf{297}\\
			& mAP$_{50}\uparrow$ &	0.5231 &	0.4030	&0.5157	&0.5137&	0.4807 & 0.4477&	\textbf{0.5861}	\\
			\hline
			\multirow{2}{*}{Panoptic Segmentation} & PQ$\uparrow$	&0.38	&0.34&	0.41	&0.39	& 0.37 & 0.23&	\textbf{0.43}	 \\
			& SQ$\uparrow$ &	0.74&	0.65&	0.76&	0.68	& 0.70 & 0.57	&\textbf{0.80}\\
			\hline
			\multirow{2}{*}{Flow Estimation}  &EPE$\downarrow$&	1.22	&1.64	&1.04	&0.93	& 0.90 & 1.72&	\textbf{0.86}\\
			& AE$\downarrow$&	20.12&	23.52&	20.56&	18.78&	15.26 & 30.62&	\textbf{10.43}\\
			\hline
			\multirow{2}{*}{Depth Estimation}  & RMSE$\downarrow$	&0.5242	&0.9012	&0.6376	&0.5654 & 0.5367	&0.7562 &	\textbf{0.4726}\\
			&  SSIM$\uparrow$	&0.8841	&0.6523	&0.8253	&0.8351	& 0.8642 &0.7345	&\textbf{0.9215}\\
			\hline\hline
		\end{tabular}
		\label{tab:application}
	\end{threeparttable}
\end{table*}

\subsubsection{Object Detection}
Object detection is a fundamental task in computer vision that involves identifying and localizing objects of interest within an image. The precision of object detection, however, is often susceptible to the visual quality of the image. For instance, in scenes with a wide dynamic range, underexposed and overexposed regions can lose significant scene details, leading to decreased detection precision. In order to demonstrate how our EventHDR method and dataset can effectively enable better object detection in scenes with challenging lighting conditions, we employ a popular object detection framework YOLOv3~\cite{redmon2018yolov3} and on different data sources for a comparative evaluation of HDR scenes.

For a thorough and comprehensive analysis, we manually annotate our EventHDR dataset established in Section \ref{sec:real_data} using the COCO~\cite{lin2014microsoft} object detection annotation format with 80 categories,  with the aid of the EISeg \cite{hao2022eiseg} labeling tool. Since the primary focus of EventHDR dataset is street views, our annotation include a broad range of vehicle types such as \textit{cars}, \textit{trucks}, and \textit{buses}, \etc. For each of the 19 test video sequences, we use frames at intervals of 20 to ensure consistency and comprehensive coverage.

\begin{figure*}[h!] \small
	\centering
	\footnotesize
	\setlength{\tabcolsep}{1pt}
	\begin{tabular}{cccccccc}
		\includegraphics[width=.122\linewidth,clip,keepaspectratio]{./figures/detection/APS.png} & 
		\includegraphics[width=.122\linewidth,clip,keepaspectratio]{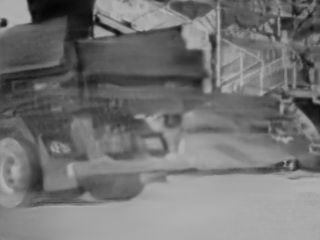} & 
		\includegraphics[width=.122\linewidth,clip,keepaspectratio]{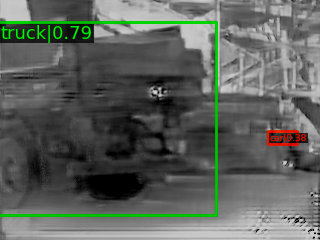} & 
		\includegraphics[width=.122\linewidth,clip,keepaspectratio]{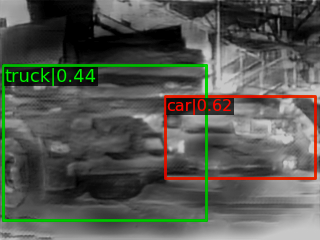} & 
		\includegraphics[width=.122\linewidth,clip,keepaspectratio]{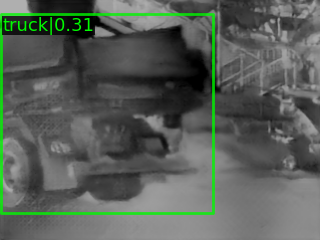} & 
		\includegraphics[width=.122\linewidth,clip,keepaspectratio]{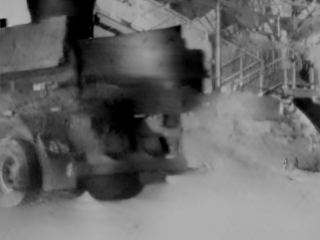} & 
		\includegraphics[width=.122\linewidth,clip,keepaspectratio]{./figures/detection/ours_real.png} & 
		\includegraphics[width=.122\linewidth,clip,keepaspectratio]{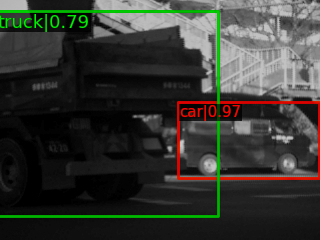} \\
		APS & E2VID~\cite{rebecq2019events} & E2VID+~\cite{stoffregen2020reducing} & EITR~\cite{weng2021event} & HyperE2V\cite{ercan2024hypere2vid} & Ours-sim & Ours & HDR 
	\end{tabular}
	\vspace{-2mm}
	\caption{\textbf{Object Detection} for HDR Scenes. For each method, we present the  reconstructed image and the corresponding object detection results using the reconstructed images as input.}
	\label{fig:application_detection}
	\vspace{-1mm}
\end{figure*}

To quantitatively evaluate each method, we provide the number of detected objects and mAP$_{50}$ score in Table~\ref{tab:application}. mAP$_{50}$ calculates at an Intersection over Union (IoU) threshold of 0.5, providing insights into the model’s capability to accurately detect objects with a moderate overlap with ground truth bounding boxes. The visual results are further provided in Fig.~\ref{fig:application_detection}. We can observe that APS suffers from severe loss of detail in both dark and bright regions, leading to the inability to detect cars in such areas. For other event-to-HDR reconstruction methods, although they attempt to recover the shape of these areas, the unsatisfactory visual quality hinders their object detection performance. As for our network trained on simulated data, the confidence level is considerably lower, which indicates that our paired HDR/event training data can help the learning process to effectively map event frames to HDR videos.

Our EventHDR method demonstrates substantial improvement in object detection performance compared to conventional APS images and other event-to-HDR methods. This is attributed to the higher visual quality of the reconstructed HDR images, which preserves crucial scene details in both dark and bright regions. As a result, the object detection framework can more accurately identify and localize objects in challenging HDR scenes. These findings not only highlight the advantages of our method and dataset but also emphasize the importance of high-quality HDR reconstruction for downstream vision tasks.

\subsubsection{Panoptic Segmentation}
Panoptic segmentation~\cite{chen2024frequency} combines the tasks of semantic segmentation and instance segmentation by jointly segmenting and classifying every pixel in an image. In this task, we apply our method to demonstrate its potential to facilitate accurate identification of both object instances and semantic classes, particularly in scenes with high dynamic range. We employ the well-established panoptic segmentation framework Mask2Former~\cite{cheng2022masked}  on the reconstructed videos of all methods, and evaluate the performance of our method against alternative methods.

\begin{figure*}[h!] \small
	\centering
	\footnotesize
	\setlength{\tabcolsep}{1pt}
	\begin{tabular}{cccccccc}
		
		\includegraphics[width=.122\linewidth,clip,keepaspectratio]{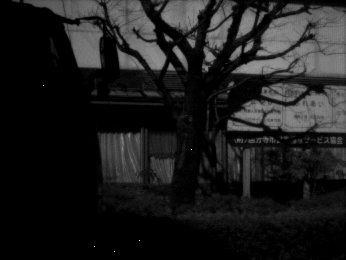} & 
		\includegraphics[width=.122\linewidth,clip,keepaspectratio]{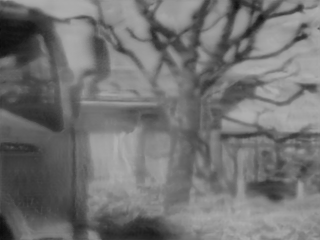} & 
		\includegraphics[width=.122\linewidth,clip,keepaspectratio]{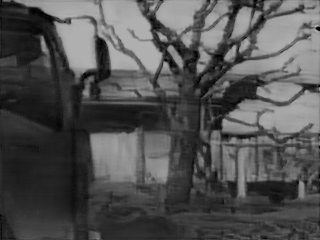} & 
		\includegraphics[width=.122\linewidth,clip,keepaspectratio]{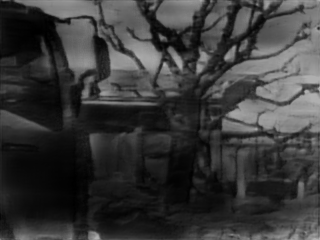} & 
		\includegraphics[width=.122\linewidth,clip,keepaspectratio]{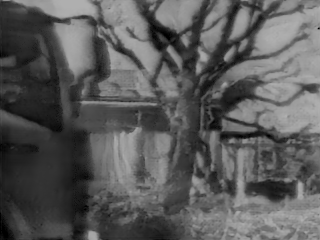} & 
		\includegraphics[width=.122\linewidth,clip,keepaspectratio]{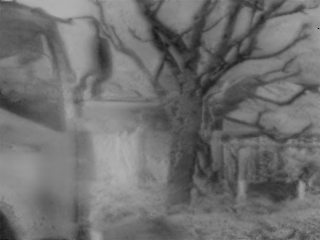} & 
		\includegraphics[width=.122\linewidth,clip,keepaspectratio]{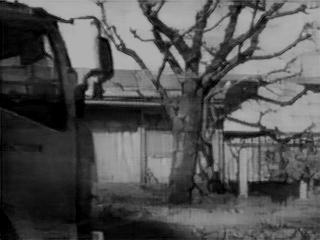} & 
		\includegraphics[width=.122\linewidth,clip,keepaspectratio]{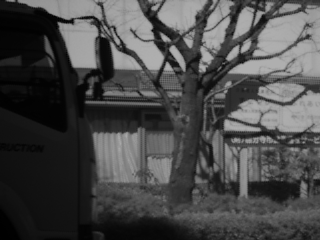}  \\
		\includegraphics[width=.122\linewidth,clip,keepaspectratio]{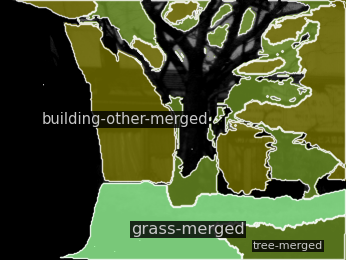} & 
		\includegraphics[width=.122\linewidth,clip,keepaspectratio]{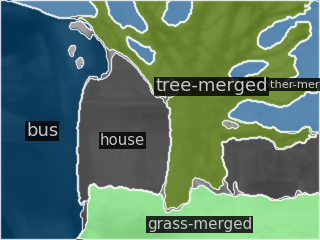} & 
		\includegraphics[width=.122\linewidth,clip,keepaspectratio]{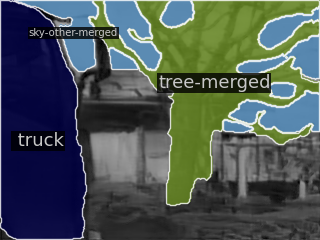} & 
		\includegraphics[width=.122\linewidth,clip,keepaspectratio]{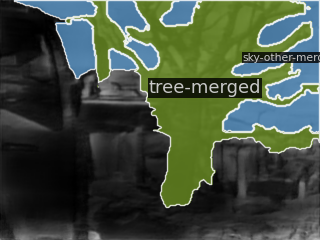} & 
		\includegraphics[width=.122\linewidth,clip,keepaspectratio]{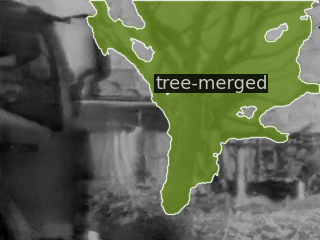} & 
		\includegraphics[width=.122\linewidth,clip,keepaspectratio]{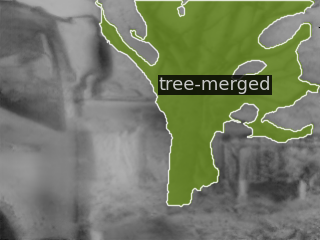} & 
		\includegraphics[width=.122\linewidth,clip,keepaspectratio]{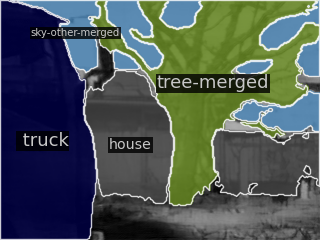} & 
		\includegraphics[width=.122\linewidth,clip,keepaspectratio]{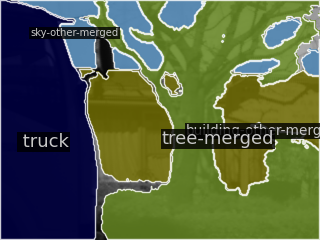} \\
		APS & E2VID~\cite{rebecq2019events} & E2VID+~\cite{stoffregen2020reducing} & EITR~\cite{weng2021event} & HyperE2V\cite{ercan2024hypere2vid} & Ours-sim & Ours & HDR 
	\end{tabular}
	\vspace{-2mm}
	\caption{\textbf{Panoptic segmentation} for HDR Scenes. The first row displays the reconstructed images for each method, while the second row presents the corresponding panoptic segmentation results, using the reconstructed images as input.}
	\label{fig:application_segmentation}
	\vspace{-2mm}
\end{figure*}

For comparison, we treat the prediction results of real HDR images as ground truth and calculate the Panoptic Quality (PQ)~\cite{kirillov2019panoptic} and Segmentation Quality (SQ) for all input data types. The results provided in Table~\ref{tab:application} and Fig.~\ref{fig:application_segmentation} show that our reconstructed HDR video leads to improved panoptic segmentation performance in real HDR scenes, demonstrating the advantages of our method in HDR panoptic segmentation.

\begin{figure*}[h!] \small
	\centering
	\footnotesize
	\setlength{\tabcolsep}{1pt}
	\begin{tabular}{cccccccc}
		
		\includegraphics[width=.122\linewidth,clip,keepaspectratio]{./figures/flow/APS-1.png} & 
		\includegraphics[width=.122\linewidth,clip,keepaspectratio]{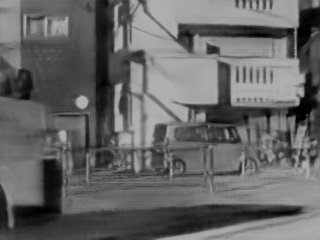} & 
		\includegraphics[width=.122\linewidth,clip,keepaspectratio]{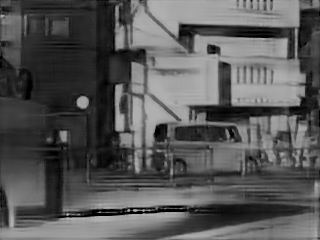} & 
		\includegraphics[width=.122\linewidth,clip,keepaspectratio]{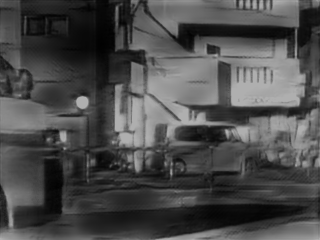} & 
		\includegraphics[width=.122\linewidth,clip,keepaspectratio]{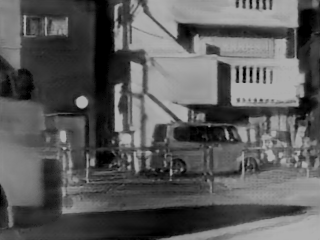} &
		\includegraphics[width=.122\linewidth,clip,keepaspectratio]{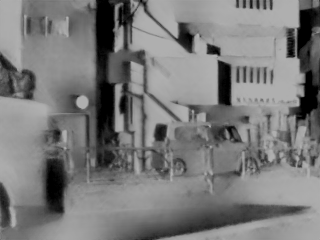} &  
		\includegraphics[width=.122\linewidth,clip,keepaspectratio]{./figures/flow/ours_real-1.png} & 
		\includegraphics[width=.122\linewidth,clip,keepaspectratio]{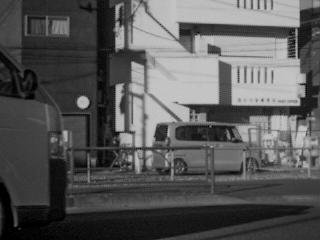}  \\
		\includegraphics[width=.122\linewidth,clip,keepaspectratio]{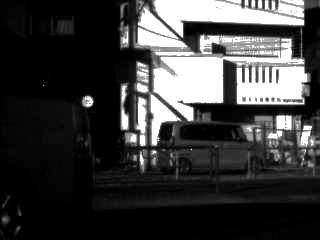} & 
		\includegraphics[width=.122\linewidth,clip,keepaspectratio]{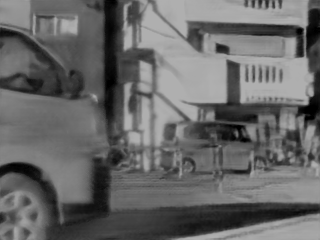} & 
		\includegraphics[width=.122\linewidth,clip,keepaspectratio]{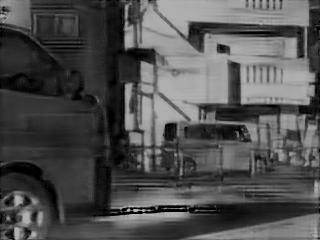} & 
		\includegraphics[width=.122\linewidth,clip,keepaspectratio]{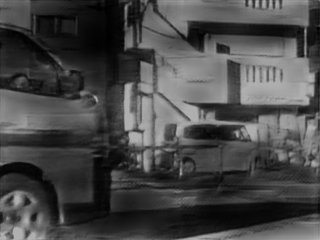} &
		\includegraphics[width=.122\linewidth,clip,keepaspectratio]{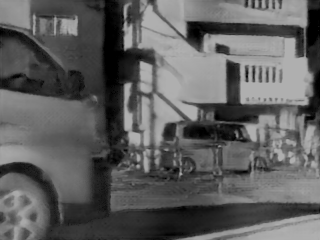} & 
		\includegraphics[width=.122\linewidth,clip,keepaspectratio]{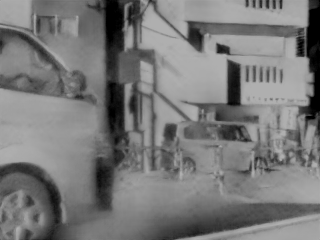} & 
		\includegraphics[width=.122\linewidth,clip,keepaspectratio]{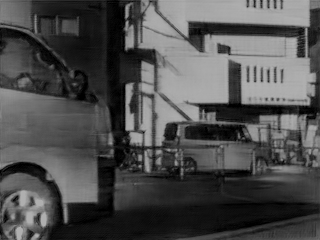} & 
		\includegraphics[width=.122\linewidth,clip,keepaspectratio]{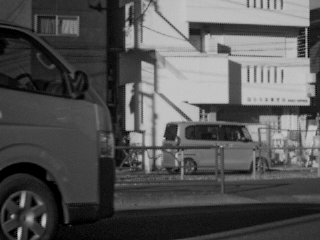}  \\
		\includegraphics[width=.122\linewidth,clip,keepaspectratio]{./figures/flow/APS-flow.png} & 
		\includegraphics[width=.122\linewidth,clip,keepaspectratio]{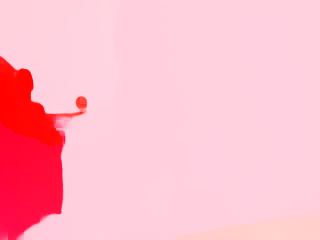} & 
		\includegraphics[width=.122\linewidth,clip,keepaspectratio]{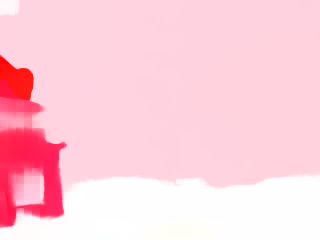} & 
		\includegraphics[width=.122\linewidth,clip,keepaspectratio]{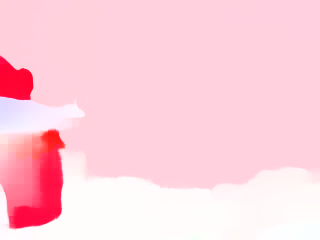} & 
		\includegraphics[width=.122\linewidth,clip,keepaspectratio]{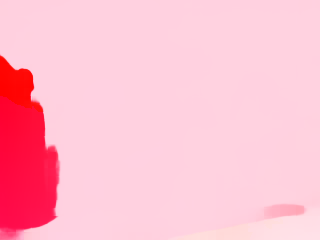} &
		\includegraphics[width=.122\linewidth,clip,keepaspectratio]{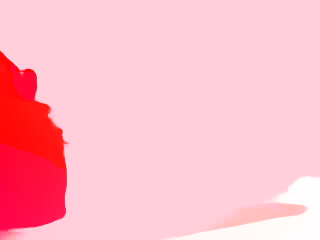} &  
		\includegraphics[width=.122\linewidth,clip,keepaspectratio]{./figures/flow/ours_real-flow.png} & 
		\includegraphics[width=.122\linewidth,clip,keepaspectratio]{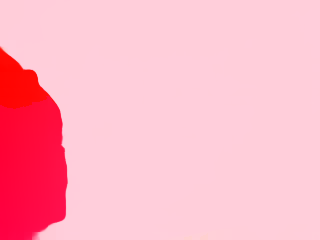}  \\
		APS & E2VID~\cite{rebecq2019events} & E2VID+~\cite{stoffregen2020reducing} & EITR~\cite{weng2021event} & HyperE2V\cite{ercan2024hypere2vid} & Ours-sim & Ours & HDR 
	\end{tabular}
	\vspace{-2mm}
	\caption{\textbf{Optical flow estimation} for HDR Scenes. The first row displays the reconstructed images for each method, while the second row presents the corresponding optical flow estimation results, using the reconstructed images as input.}
	\label{fig:application_flow}
	\vspace{-1mm}
\end{figure*}

\subsubsection{Optical Flow Estimation}
Optical flow estimation is the process of estimating the motion of objects in a scene by analyzing the pattern of apparent motion. This task is crucial for applications such as video stabilization, object tracking~\cite{wang2024multi}, and action recognition. However, it becomes challenging when motion is difficult to recognize under extreme HDR conditions. We apply our method to optical flow estimation to verify its capacity for  reconstructing fast-moving scenes with varying lighting conditions. The widely used RAFT~\cite{teed2020raft} algorithm is used to estimate the optical flow for all data sources.

Considering that tone-mapped ground truth HDR images retain most of the scene details, we further use the optical flow predicted by these tone-mapped images as ground truth optical flow to measure the precision of other methods quantitatively. Regarding the metric employed for this evaluation, we use the end-point error (EPE) and	Angular Error (AE). Among them, EPE quantifies the difference between the estimated optical flow and the actual motion, and AE is useful for emphasizing the directional accuracy of the flow estimation. By comparing our method with alternative approaches in Table~\ref{tab:application} and Fig.~\ref{fig:application_flow}, we demonstrate that our reconstructed HDR video provides superior optical flow estimation performance, owing to the successful recovery of the darkest and brightest regions in HDR scenes. This underlines the benefits of our approach in handling dynamic scenes with challenging lighting conditions.

\begin{figure*}[h!] \small
	\centering
	\footnotesize
	\setlength{\tabcolsep}{1pt}
	\begin{tabular}{ccccccccc}
		\includegraphics[width=.122\linewidth,clip,keepaspectratio]{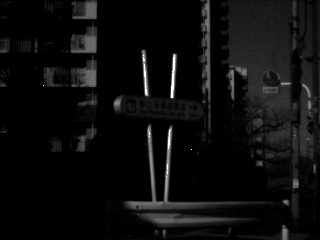} & 
		\includegraphics[width=.122\linewidth,clip,keepaspectratio]{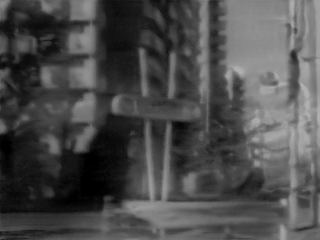} & 
		\includegraphics[width=.122\linewidth,clip,keepaspectratio]{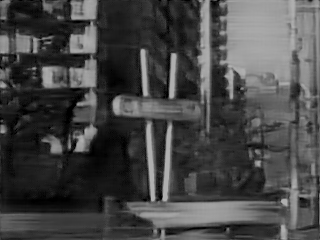} & 
		\includegraphics[width=.122\linewidth,clip,keepaspectratio]{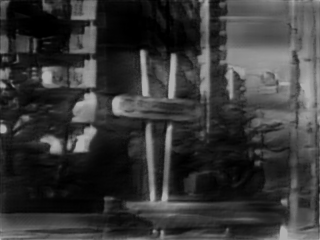} & 
		\includegraphics[width=.122\linewidth,clip,keepaspectratio]{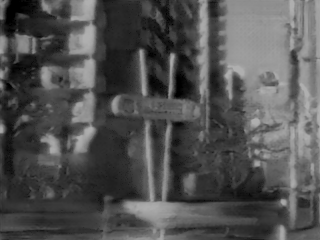} & 
		\includegraphics[width=.122\linewidth,clip,keepaspectratio]{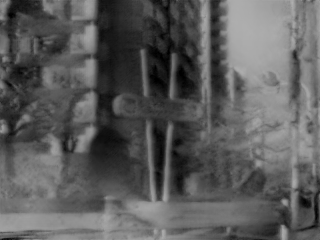} & 
		\includegraphics[width=.122\linewidth,clip,keepaspectratio]{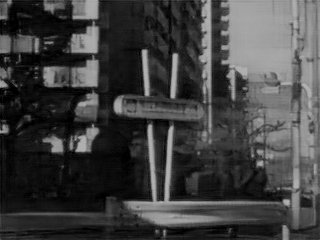} & 
		\includegraphics[width=.122\linewidth,clip,keepaspectratio]{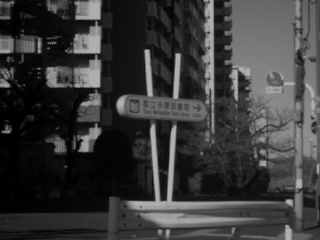}  \\
		\includegraphics[width=.122\linewidth,clip,keepaspectratio]{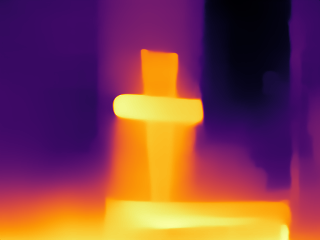} & 
		\includegraphics[width=.122\linewidth,clip,keepaspectratio]{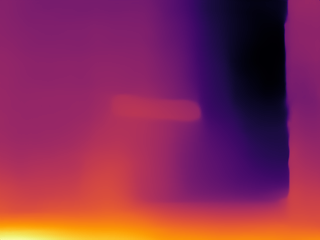} & 
		\includegraphics[width=.122\linewidth,clip,keepaspectratio]{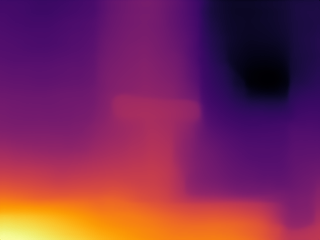} & 
		\includegraphics[width=.122\linewidth,clip,keepaspectratio]{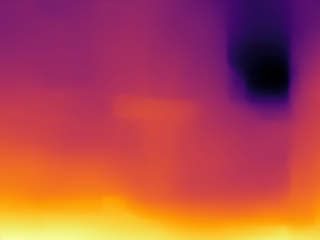} & 
		\includegraphics[width=.122\linewidth,clip,keepaspectratio]{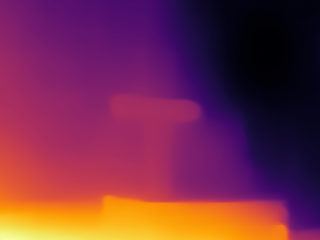} & 
		\includegraphics[width=.122\linewidth,clip,keepaspectratio]{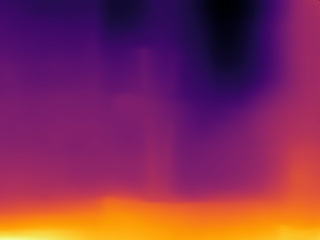} & 
		\includegraphics[width=.122\linewidth,clip,keepaspectratio]{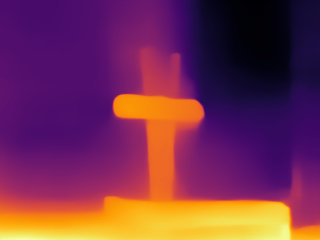} & 
		\includegraphics[width=.122\linewidth,clip,keepaspectratio]{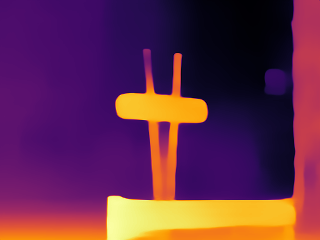} \\
		APS & E2VID~\cite{rebecq2019events} & E2VID+~\cite{stoffregen2020reducing} & EITR~\cite{weng2021event} & HyperE2V\cite{ercan2024hypere2vid} & Ours-sim & Ours & HDR 
	\end{tabular}
	\vspace{-2mm}
	\caption{\textbf{Monocular depth estimation} for HDR Scenes. The first and second rows display two neighboring  reconstructed frames for each method, while the third row presents the corresponding monocular depth estimation results, using the reconstructed images as input.}
	\label{fig:application_depth}
	\vspace{-2mm}
\end{figure*}

\subsubsection{Monocular Depth Estimation}
Monocular depth estimation is the task of estimating scene depth from a single image. It is an essential task for applications such as autonomous navigation, 3D reconstruction, and augmented reality. We evaluate our method on this task to demonstrate its ability to reconstruct HDR video that preserves depth information even under challenging illumination conditions. Using a state-of-the-art monocular depth estimation framework MiDaS~\cite{ranftl2020towards}, we compare the performance of our method against alternative data sources.

Additionally, the prediction results of real HDR images are used as ground truth to assess the results of other methods, and we calculate the root-mean-square-error (RMSE) for numerical evaluation, as well as SSIM  to evaluate the structural accuracy of the predicted depth map.  The numeric and visual results are shown in Table~\ref{tab:application} and Fig.~\ref{fig:application_depth}. Our results show that our reconstructed HDR video leads to improved depth estimation accuracy, further emphasizing the effectiveness of our method for downstream vision tasks in challenging lighting conditions.

\subsubsection{Discussion}
The experiments above indicate that our method and dataset show robust performance across essential vision tasks by effectively leveraging event cameras' HDR capabilities. This highlights the potential of our approach  to enhance computer vision applications especially in variable lighting and dynamic scenarios.

\begin{figure}[t] \small
	\centering
	\setlength{\tabcolsep}{1pt}
	\begin{tabular}{cccc}
		\includegraphics[width=.24\linewidth,clip,keepaspectratio]{./figures/depth/ref.png} & 
		\includegraphics[width=.24\linewidth,clip,keepaspectratio]{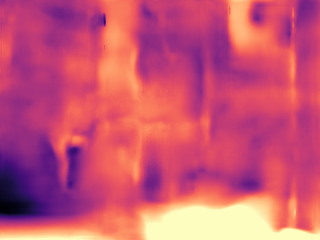} &
		\includegraphics[width=.24\linewidth,clip,keepaspectratio]{./figures/depth/ours_real-depth.png} & 
		\includegraphics[width=.24\linewidth,clip,keepaspectratio]{./figures/depth/ref-depth.png} \\
		HDR Image & E2Depth\cite{hidalgo2020learning} & Our Depth & HDR Depth
	\end{tabular}
	\vspace{-1mm}
	\caption{Comparison between event-based and two-stage methods.}
	\vspace{-2mm}
	\label{fig:discussion}
\end{figure}

Besides the two-stage manner that transforms event streams to intensity images and then performs frame-based downstream approaches, there are some event-based methods that process event streams directly. Theoretically, event-based methods are expected to show superiority in scenarios where they are trained on specific, well-labeled datasets, achieving high accuracy by tailoring their models to the exact conditions of those datasets. However, this specialization often reduces their generalizability, leading to suboptimal performance on more diverse, general event data. Examples of comparing with event-based depth estimation method E2Depth~\cite{hidalgo2020learning} is shown in Fig.~\ref{fig:discussion}. In contrast, by reconstructing HDR images from event data and applying existing pretrained vision models, our method circumvents the need for task-specific training or model adaptation. This not only simplifies implementation but also enables the seamless integration of HDR reconstructions into a wide range of vision tasks, making the frame-based approach more practical and flexible for diverse applications.

\section{Conclusion}
To leverage both the high-speed and HDR capabilities of event cameras, as well as well-established frame-based computer vision algorithms, we present a novel recurrent convolutional neural network for reconstructing standard HDR videos from event streams. Specifically, our model consists of a key frame guided recurrent feature extractor, which exploits features from a long temporal range while discarding long-term error accumulation. Then, we introduce a deformable convolution-based feature alignment module, a fusion module with attention mechanism to reconstruct high-quality HDR videos. We also employ a temporal consistency loss to minimize discrepancies between the reconstructions and real-world scenes. Furthermore, we design a customized imaging system to capture synchronous event and HDR data, providing the first dataset with paired high-speed HDR and event data of real dynamic scenes. Experimental results have verified the effectiveness of our proposed high-speed HDR video reconstruction method and our collected paired EventHDR dataset.

Our novel co-axis imaging system paves the way for the reconstruction of high bit-depth HDR formats through events. Thanks to the high-quality paired real data, we tackle the long-standing problem of unreal artifacts in the high-speed event-to-HDR reconstruction task. We believe this co-axis imaging system has broader potential beyond the event-to-HDR task, such as creating real paired datasets for event-guided video interpolation. In future work, we would like to further investigate the applications of our imaging system to build a more systematic and comprehensive event processing pipeline.


%
%
%
%
%

\ifCLASSOPTIONcaptionsoff
  \newpage
\fi



\bibliographystyle{IEEEtran}
\bibliography{IEEEabrv,bibtex/egbib.bib}
%
%
%

%

\begin{IEEEbiography}[{\includegraphics[width=1in,height=1.25in,clip,keepaspectratio]{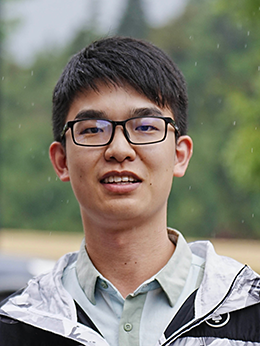}}]{Yunhao Zou}
	received the B.S. degree of Computer Science from Beijing Institute of Technology in 2019. He is currently pursing the Ph.D. degree of Computer Science in Beijing Institute of Technology. His research interests include low-level vision, computational photography and computational imaging. He serves as reviewer for major international conferences and journals, including TPAMI, CVPR, ICCV, ECCV, ICLR, AAAI, BMVC, \etc.
\end{IEEEbiography}

\begin{IEEEbiography}[{\includegraphics[width=1in,height=1.25in,clip,keepaspectratio]{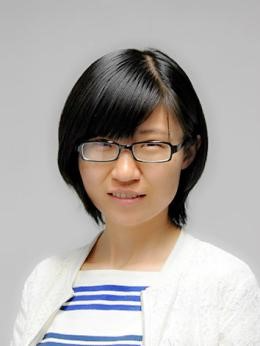}}]{Ying Fu}
	(Senior Member, IEEE) received the B.S. degree in electronic engineering from Xidian University, Xi’an, China, in 2009, the M.S. degree in automation from Tsinghua University, Beijing, China, in 2012, and the Ph.D. degree in information science and technology from the University of Tokyo, Tokyo, Japan, in 2015. She is currently a Professor with the School of Computer Science and Technology, Beijing Institute of Technology. Her research interests include physics-based vision, image and video processing, and computational photography.
\end{IEEEbiography}

\begin{IEEEbiography}[{\includegraphics[width=1in,height=1.25in,clip,keepaspectratio]{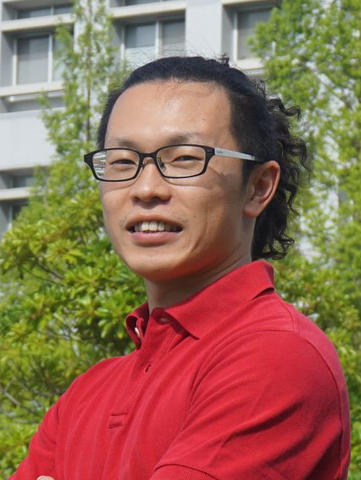}}]{Tsuyoshi Takatani}
	received his doctoral degree from Nara Institute of Science and Technology (NAIST) in 2019. He is currently an assistant professor of the Institute of Systems and Information Engineering at University of Tsukuba. He leads the Computational Imaging and Graphics Laboratory as the founding director. His research interests include computational imaging and fabrication, and inverse rendering. He is a member of the IEEE and the OPTICA.
\end{IEEEbiography}

\begin{IEEEbiography}
	[{\includegraphics[width=1in,height=1.25in,clip,keepaspectratio]{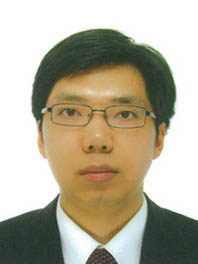}}]{Yinqiang Zheng} is currently a full professor in the Next Generation Artificial Intelligence Research Center, The University of Tokyo. He received a Doctoral degree of engineering from the Department of Mechanical and Control Engineering, Tokyo Institute of Technology, Tokyo, Japan, in 2013. His research interests include optical imaging, computer vision and artificial intelligence. He received the Konica Minolta Image Science Award and Funai Academic Award. He is a Senior Member of IEEE.
\end{IEEEbiography}



\end{document}